%% file: R1-Aug-2021.tex
\documentclass[10pt,journal,compsoc]{IEEEtran}

\usepackage{graphicx}

\usepackage{float}
\usepackage{graphicx}
\usepackage{comment}
\usepackage{amsmath,amssymb} %
\usepackage{color}
\usepackage{xspace}
\usepackage{textcomp}
\usepackage{cite}
\usepackage{mathtools}
\usepackage{bbm}
\usepackage{makecell, verbatim, sidecap}
\usepackage{subfig}
\usepackage{multirow}
\usepackage{url,nicefrac}

\usepackage[margin=4pt,font=small,labelfont=bf,labelsep=endash,tableposition=bottom]{caption}

\renewcommand{\vec}[1]{\ensuremath{\pmb{#1}}}
\newcommand{\mat}[1]{\ensuremath{\mathbf{#1}}}
\newcommand{\set}[1]{\ensuremath{\mathscr{#1}}}

\makeatletter
\@tfor\next:=abcdefghijklmnopqrstuvwxyxz\do
{\begingroup\edef\x{\endgroup
		\noexpand\@namedef{v\next}{\noexpand\vec{\next}}%
	}\x}
\@tfor\next:=ABCDEFGHIJKLMNOPQRSTUVWXYZ\do
{\begingroup\edef\x{\endgroup
		\noexpand\@namedef{m\next}{\noexpand\mat{\next}}%
	}\x}
\@tfor\next:=ABCDEFGHIJKLMNOPQRSTUVWXYZ\do
{\begingroup\edef\x{\endgroup
		\noexpand\@namedef{s\next}{\noexpand\set{\next}}%
	}\x}
\makeatother

\def\eg{{\it e.g.}\xspace}
\def\ie{{\it i.e.}\xspace}

\def\R{{\mathbb R}}

\def\pos{{\rm pos}}

\def\sigmoid{{\rm sigmoid}}
\def\ReLU{{\rm ReLU}}
\def\centerness{ {  \rm centerness  } }

\DeclarePairedDelimiter{\floor}{\lfloor}{\rfloor}

\usepackage{hyperref}
\hypersetup{colorlinks=true,breaklinks=true, urlcolor= blue,linkcolor= red,citecolor=blue
}

\def\Ours{{CondInst}}

\newcommand{\etal}{\textit{et al.}}

\usepackage{color}

\def\softmax  {{\rm softmax}}

\newcommand{\code}[1]{{\ensuremath{\tt #1}}}

\let \texttt    \code

\begin{document}

\title{Instance and Panoptic Segmentation Using Conditional Convolutions}

\author{
         Zhi Tian,  ~~~ 
         Bowen Zhang, ~~~
         Hao Chen,  ~~~
         Chunhua Shen
\thanks{
	{\it 
	Accepted to IEEE Trans.\ on Pattern Analysis and Machine Intelligence (TPAMI),
	20 Jan. 2022.
}\protect\\
            Work was done when all authors were with
            The University of Adelaide, Australia.i 
            %
			B. Zhang is with The University of Adelaide.
            Z. Tian is with Meituan Inc.
            H. Chen 
            and 
            C. Shen are  with Zhejiang  University, China.
            \protect\\
            C. Shen is the
            corresponding author  (e-mail: \code{chunhua@me.com}).
        }
}

\IEEEtitleabstractindextext{
\begin{abstract}
We propose a simple yet effective framework for instance and panoptic segmentation, termed CondInst (conditional convolutions for instance and panoptic segmentation). In the literature, 
top-performing instance segmentation methods %
typically follow the paradigm of 
Mask R-CNN and rely on ROI operations (typically ROIAlign) to attend to each instance. In contrast, we propose to attend to the instances with dynamic conditional convolutions. Instead of using instance-wise ROIs as inputs to the instance mask head of fixed weights, we 
design 
dynamic instance-aware mask heads, conditioned on the instances to be predicted. CondInst enjoys three advantages: 1) Instance and panoptic segmentation are unified into a fully convolutional network, eliminating the need for ROI cropping and feature alignment. 2) The elimination of the ROI cropping also significantly improves the output instance mask resolution. 3) Due to the much improved capacity of dynamically-generated conditional convolutions, the mask head can be very compact (\eg, 3 conv.\ layers, each having only 8 channels), leading to significantly %
faster 
inference time per instance and making the overall inference time less relevant to 
the number of instances. We demonstrate a simpler method that can achieve improved accuracy and inference speed on both instance and panoptic segmentation tasks. On the COCO dataset, we outperform a few state-of-the-art methods. We hope that 
CondInst can be a strong baseline for instance and panoptic segmentation. 
Code is available at:
\def\UrlFont{\sf \color{blue}}
\url{https://git.io/AdelaiDet}

\end{abstract}


}
 

\maketitle

\begin{figure}[t]
\centering 
\includegraphics[width=.45\textwidth]{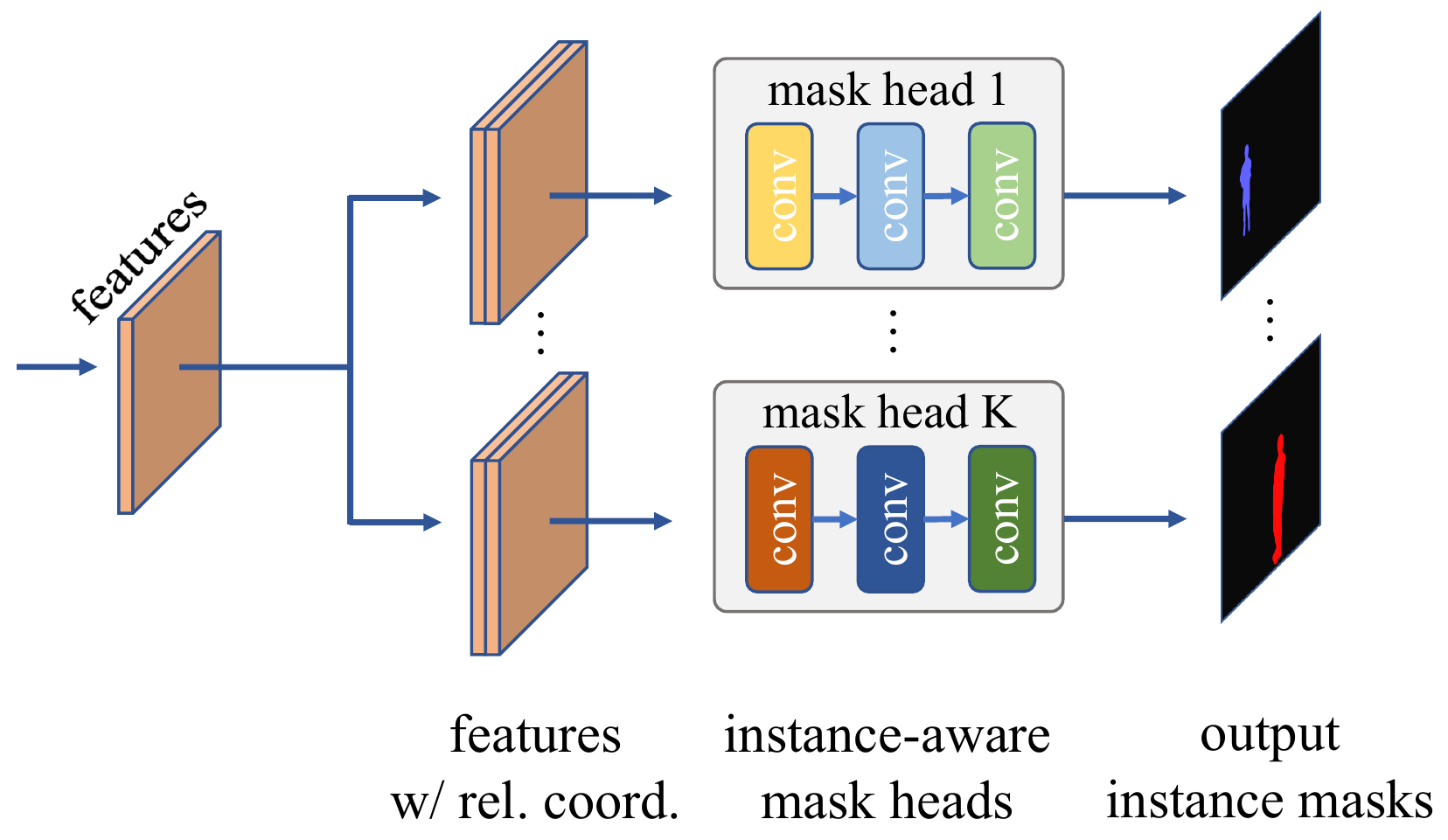}
\caption{\Ours\ uses instance-aware mask heads to predict the mask for each instance. $K$ is the number of instances to be predicted. Note that each output map only contains the mask of one instance. The filters in the mask head vary with different instances, which are dynamically-generated and conditioned on the target instance. \ReLU\ is used as the activation function (excluding the last conv.\ layer).}
\label{fig:mask_heads}
\end{figure}

\section{Introduction}
\begin{figure*}[t]
	\centering
	\includegraphics[width=.9\linewidth]{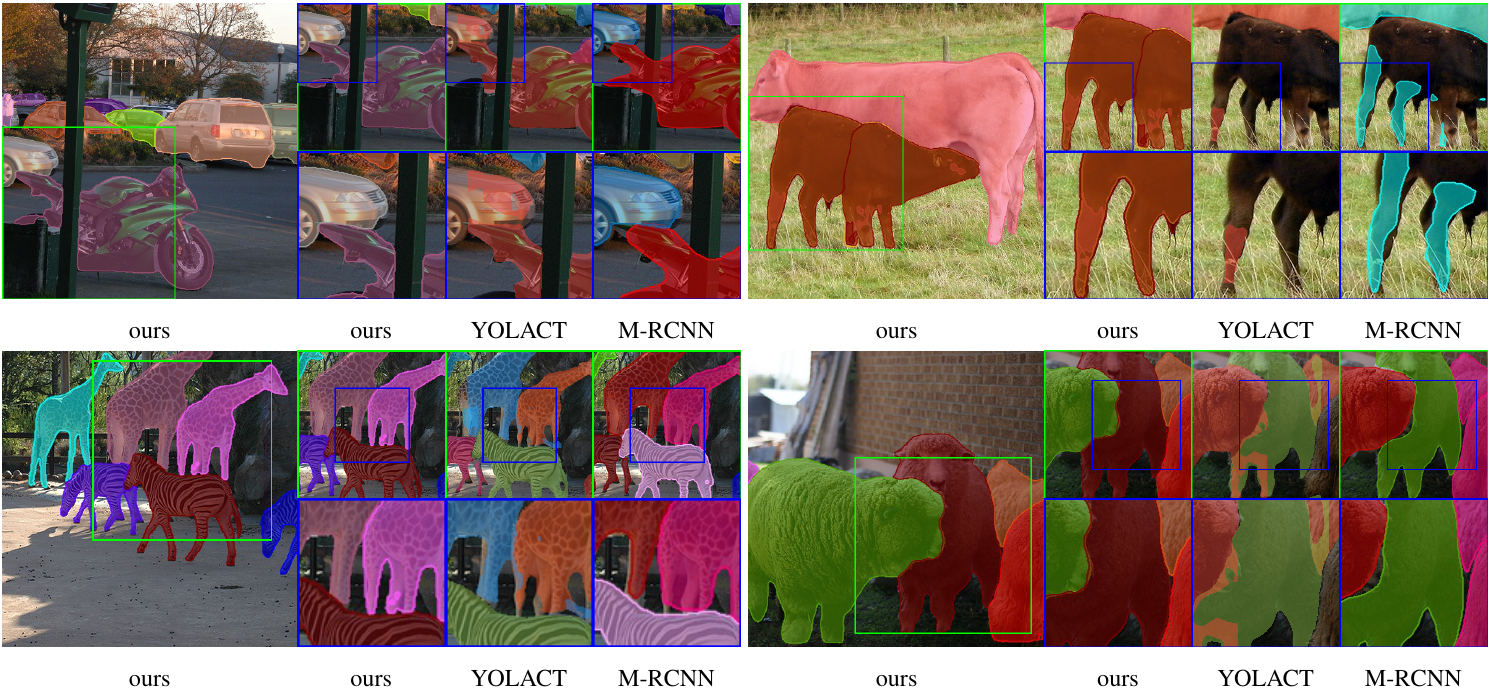}
	\caption{Qualitative comparisons with other methods. We compare the proposed \Ours\ against YOLACT~\cite{bolya2019yolact} and Mask R-CNN~\cite{he2017mask}. Our masks are generally of higher quality (\eg, preserving %
	finer 
	details). Best viewed on screen.}
	\label{fig:qualitative}
\end{figure*}

Instance segmentation is a fundamental yet challenging task in computer vision, which requires
an
algorithm to predict a per-pixel mask with a category label for each instance of interest in an image. Panoptic segmentation further requires the algorithm to segment the stuff (\eg, sky and grass), %
assigning 
every pixel in the image a semantic label. Panoptic segmentation is often built on an instance segmentation framework with an extra semantic segmentation branch. Therefore, both instance and panoptic segmentation share the same key challenge----how to efficiently and effectively distinguish individual instances.

Despite a few works being proposed recently,
the dominant method tackling this challenge is still the two-stage
method such as Mask R-CNN \cite{he2017mask}, which casts instance segmentation into a two-stage detection-and-segmentation task. To be specific, Mask R-CNN first employs an object detector Faster R-CNN
to predict a bounding-box  for  each instance.
Then for each instance, regions-of-interest (ROIs) are cropped from the networks' feature maps using the ROIAlign operation.
To predict the final masks for each instance, a compact
fully convolutional network (FCN) (\ie, mask head) is applied to these ROIs
to perform foreground/background segmentation.
However, this ROI-based method may have
the following drawbacks. 1) Since
ROIs are often axis-aligned bounding-boxes, for
objects with irregular shapes, they may contain an excessive amount of irrelevant image content including background and other instances.
This issue may be
mitigated by
using
rotated RoIs~\cite{wu2019detectron2},
but with the price of a more complex pipeline.
2) In order to distinguish between the foreground instance and the background stuff or instance(s) with the fixed mask head, the mask head needs a strong capacity and a relatively larger receptive field to encode
sufficiently
large context information.
As a result, a stack of $3 \times 3$ convolutions is used in the mask head (\eg, four $3 \times 3$ convolutions with $256$ channels in Mask R-CNN).
It
considerably
increases
computational complexity
of the mask head, resulting that the inference time significantly varies
in
the number of instances.
3)
ROIs are typically
of
different sizes. In order to use effective batched computation in modern deep learning frameworks \cite{pytorch, tensorflow}, a resizing operation is often required to resize the cropped regions into patches
of the same size.
For instance, Mask R-CNN resizes all the cropped regions to $14 \times 14$ (upsampled to $28 \times 28$ using a deconvolution), which
restricts
the output resolution of instance segmentation, as
large instances would require
higher resolutions to retain details at the boundary.

In computer vision,
the
closest
task to instance segmentation is semantic segmentation, for which fully convolutional networks (FCNs) have shown dramatic success \cite{long2015fully, chen2017deeplab, tian2019decoders, he2019knowledge, liu2020structured}. FCNs also have shown excellent performance on many other
per-pixel prediction
tasks
ranging from low-level image processing such as denoising, super-resolution;
to mid-level tasks such as optical flow estimation and contour detection;
and high-level tasks including
recent single-shot object detection \cite{tian2019fcos}, monocular depth estimation \cite{Depth2015Liu, Yin2019enforcing, %
bian2019unsupervised
} and counting \cite{boominathan2016crowdnet}.
However,
almost all the instance segmentation methods based on FCNs\footnote{By FCNs, we mean the
	vanilla
	FCNs in \cite{long2015fully} that
	only
	involve convolutions and pooling.
} lag behind state-of-the-art ROI-based methods.
{Why do the versatile FCNs perform unsatisfactorily on instance segmentation?}
This is due to the fact that the FCNs tend to yield similar predictions for similar image appearance. As a result, the vanilla FCNs are incapable of distinguishing individual instances. For example, if two persons {\tt A} and {\tt B} with the similar appearance are in an input image, when predicting the instance mask of {\tt A},
the FCN needs to predict
{\tt B}
as background \textit{w.r.t.}\
{\tt A}, which can be difficult as they
look similar in 
appearance. Therefore, an ROI operation is %
used
to crop the person of interest, \ie, {\tt A}; and
filter out {\tt B}. Essentially, this is the core operation making the model attend to an instance.

\begin{figure*}[t]
	\centering
	\includegraphics[width=.769\linewidth]
	{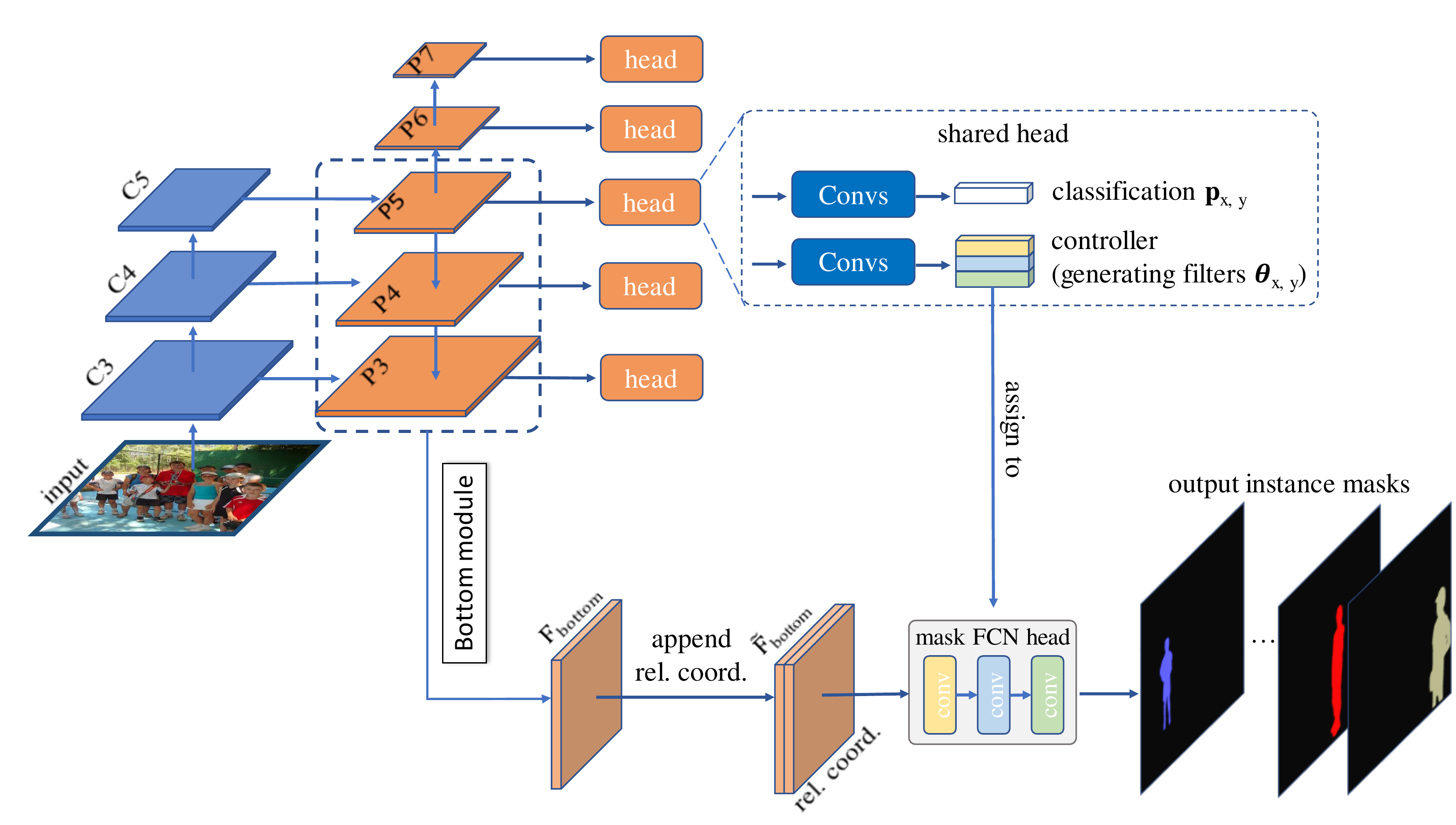}
	\caption{The overall architecture of  \textbf{\Ours}. 
	$C_3$, $C_4$ and $C_5$ are the feature maps of the backbone network (\eg, ResNet-50). $P_3$ to $P_7$ are the FPN feature maps as in \cite{lin2017feature, tian2019fcos}. $\mF_{bottom}$ is the bottom branch's output, whose resolution is the same as that of $P_3$. Following~\cite{chen2020blendmask}, the bottom branch aggregates the feature maps $P_3$, $P_4$ and $P_5$. $\tilde{\mF}_{bottom}$ is obtained by concatenating  the relative coordinates to $\mF_{bottom}$. The classification head predicts the class probability $\vp_{x, y}$ of the target instance at location $(x, y)$, same as in FCOS. The controller generates the filter parameters $\vec{\theta}_{x, y}$ of the mask head for the instance. Similar to FCOS, there are also center-ness and box heads in parallel with the controller (not shown in the figure for simplicity). Note that the heads in the dashed box are repeatedly applied to  $P_3$ $\cdots$$P_7$. 
	The mask head is instance-aware, and is applied to $\tilde{\mF}_{bottom}$ as many times as the number of instances in the image (refer to Fig.~\ref{fig:mask_heads}).
	}
	\label{fig:main_figure}
\end{figure*}

In this work, we advocate a new solution for instance segmentation, termed \Ours. Instead of using ROIs, \Ours\ attends to each instance by
using \textit{instance-sensitive convolution filters} 
as well as relative coordinates that are appended to the feature maps. Specifically, unlike Mask R-CNN, which uses a standard convolution network with a set of fixed convolutional filters as the mask head for predicting all instances, the network parameters in our mask head are adapted
according to the instance to be predicted. Inspired by dynamic filtering networks \cite{jia2016dynamic} and CondConv \cite{yang2019condconv}, for each instance,
a controller sub-network (see Fig.~\ref{fig:main_figure})
dynamically generates the mask head's filters
(conditioned on the center area of the instance), which is then used to predict the mask of this instance. It is expected that the network parameters can encode the characteristics (\eg, relative position, shape and appearance) of this instance, and only fires on the pixels of this instance, which thus bypasses the difficulty in the standard FCNs.
These conditional mask heads are applied to the whole high-resolution feature maps, \textit{thus eliminating the need for ROI operations}. At the first glance, the idea may
not work well
as instance-wise mask heads
may
incur a large number of network parameters provided that some images contain as many as dozens of instances. However, as the mask head filters are only asked to predict the mask of only one instance, it largely eases the learning requirement and thus reduces the load of the filters. As a result, the mask head can be extremely light-weight.
We will show
that %
a
very compact mask head with dynamically-generated filters can already outperform previous ROI-based Mask R-CNN. This compact mask head also results in much reduced computational complexity per instance than that of the mask head in Mask R-CNN.

We summarize our main contributions as follow.
\begin{itemize}
	\itemsep 0cm
	\item
	We attempt to solve instance segmentation
	from a new perspective that uses dynamic mask heads. This novel solution achieves improved instance segmentation performance than existing methods such as Mask R-CNN. To our knowledge, this is the first time that a new instance segmentation framework outperforms recent state-of-the-art both in accuracy and speed.
	
	\item \Ours\ is fully convolutional and avoids the aforementioned resizing operation used in many existing methods, as \Ours\ does not rely on ROI operations. Without having to resize feature maps leads to high-resolution instance masks with more accurate edges, as shown in Fig.~\ref{fig:qualitative}.

	\item Since the mask head in \Ours\ is very compact and light-weight, compared with the box detector FCOS, \Ours\ needs only $\sim$10\% more computational time (less than 5 milliseconds) to obtain the mask results of all the instances, even when processing the maximum number of instances per image (\ie, $100$ instances). As a result, the overall inference time is stable as it almost does not depend on the number of instances in the image.

	\item With an extra semantic segmentation branch, \Ours\ can be easily extended to panoptic segmentation~\cite{PanoSeg}, resulting a unified fully convolutional network for both instance and panoptic segmentation tasks.
	
	\item \Ours\ achieves state-of-the-art performance on both instance and panoptic segmentation tasks while being fast and simple. We hope that \Ours\ can be a new strong alternative for instance and panoptic segmentation tasks, as well as other instance-level recognition tasks such as keypoint detection.
	
\end{itemize}

\section{Related Work}
Here we review some work that is most relevant to ours.

\noindent\textbf{Conditional Convolutions/Dynamic filters.} Unlike traditional convolutional layers, which have fixed filters once trained, the filters of conditional convolutions are conditioned on the input and are dynamically generated by another network (\ie, a controller). This idea has been
explored previously
in dynamic filter networks \cite{jia2016dynamic} and CondConv \cite{yang2019condconv} mainly for the purpose
of
increasing  the capacity of a classification network. DGMN \cite{Zhang_2020_CVPR} also employs dynamic filters to generate the node-specific filters for message calculation, which improves the  capacity of the networks and thus results in better performance. In this work, we
extend
this idea to generate the mask head's filters conditioned on each instance, and present a high-performance instance segmentation method without the need for ROIs.

\noindent\textbf{Instance Segmentation.}
To date,
the dominant framework for  instance segmentation is still Mask R-CNN. Mask R-CNN first employs an object detector to detect the bounding-boxes of instances (\eg, ROIs). With these bounding-boxes, an ROI operation is used to crop the features of the instance from the feature maps. Finally, an
FCN head is
used
to obtain the desired instance masks. Many works \cite{chen2019hybrid, liu2018path, huang2019mask} with top performance are built on Mask R-CNN.
Moreover, some works have explored to apply the standard FCNs~\cite{long2015fully} to instance segmentation.
InstanceFCN \cite{dai2016instance} may be the first instance segmentation method
that
is fully convolutional. InstanceFCN proposes to predict position-sensitive score maps with vanilla FCNs. Afterwards, these score maps are assembled to obtain the desired instance masks.
Note that
InstanceFCN does not work well with overlapping instances.
Others
\cite{neven2019instance, newell2017associative, fathi2017semantic} attempt to first
perform
image segmentation and then the desired instance masks are formed by assembling the pixels of the same instance. Deep Watershed~\cite{bai2017deep} models instance segmentation with classical watershed transform, and object instances can be viewed as the energy basins in the energy map of the watershed transform of an image. SGN~\cite{liu2017sgn} uses a sequence of networks to gradually group the raw pixels to line segments, connected components, and finally object instances, achieving impressive performance. The single-shot Box2Pix~\cite{uhrig2018box2pix} solves instance segmentation in the bottom-up fashion.
Novotny \textit{et al.}\ \cite{Novotny_2018_ECCV} propose semi-convolutional operators to make FCNs applicable to instance segmentation. Arnab \textit{et al.}~\cite{arnab2017pixelwise} propose dynamically instantiated CRF (Conditional Random Field) for instance segmentation, which is able to produce a variable number of instances per image.
To our knowledge, thus far
none of
these
methods can outperform Mask R-CNN both in accuracy and speed on
the
COCO benchmark dataset.

The recent YOLACT \cite{bolya2019yolact}
and BlendMask \cite{chen2020blendmask} may be viewed as a reformulation of Mask RCNN, which decouples ROI detection and
feature maps used for mask prediction.
Wang \textit{et al.}\ developed a simple FCN based instance segmentation method, which segments the instances by the their locations, showing competitive performance \cite{wang2019solo, wang2020solov2}. PolarMask~\cite{polarmask} developed a new simple mask
representation for instance segmentation,
which  extends  the bounding box detector FCOS~\cite{tian2019fcos}.

\noindent\textbf{Panoptic segmentation.} There are two %
main 
approaches %
for solving 
this task. The first one is the bottom-up approach.
It tackles the task as a semantic segmentation at first and then uses clustering/grouping methods to assemble the pixels into individual instances or stuff~\cite{yang2019deeperlab, li2018weakly}. The authors of \cite{li2018weakly} also explore the weakly- or semi-supervised panoptic segmentation. The second 
approach is the %
top-down approach, which is often built on top-down instance segmentation methods. For example, Panoptic-FPN~\cite{PanoFPN} extends an additional semantic segmentation branch from Mask R-CNN %
and combines  the results with the instance segmentation results generated by Mask R-CNN~\cite{PanoSeg}. Moreover, attention based methods recently gain much popularity in many computer vision tasks, which provide a new approach to panoptic segmentation. Axial DeepLab \cite{wang2020axial} used a carefully designed module to enable attention to be applied to large-size images for panoptic segmentation.
\Ours\ can easily be applied to panoptic segmentation following the top-down approaches. We empirically observe that the quality of the instance segmentation results may be the dominant factor to the final performance.
Thus in \Ours, without bells and whistles, by simply applying the same method used by Panoptic-FPN, the panoptic segmentation performance of \Ours\ is already competitive compared to the state-of-the-art panoptic segmentation methods.

Additionally, AdaptIS \cite{sofiiuk2019adaptis} recently proposes to solve panoptic segmentation with FiLM \cite{perez2018film}. The idea shares some similarity with \Ours\, in which information about an instance is encoded in the coefficients generated by FiLM.
Since only the batch normalization coefficients are dynamically generated, AdaptIS needs a large mask head to achieve good performance. In contrast, \Ours\ directly encodes them into the conv.\  filters of the mask head, which is much more straightforward and efficient.
Also, as shown in experiments, \Ours\ can achieve much better panoptic segmentation accuracy than AdaptIS, which suggests that \Ours\ is much more effective.

\section{Our Methods: Instance and Panoptic Segmentation with \Ours}
We first present \Ours\ for instance segmentation, and then we show how the instance segmentation framework can be easily extended to panoptic segmentation by using a new semantic branch.
\subsection{Overall Architecture for Instance Segmentation}
Given an input image $I \in \R^{H \times W \times 3}$, the goal of instance segmentation is to predict the pixel-level mask and the category of each instance of interest in the image. The ground-truths are defined as $\{(\mM_i, c_i)\}$, where $\mM_i \in \{0, 1\}^{H \times W}$ is the mask for the $i$-th instance and $c_i \in \{1, 2, ..., C\}$ is the category. $C$ is $80$ on MS-COCO \cite{lin2014microsoft}. In semantic segmentation, the prediction target of each pixel are well-defined, which is the semantic category of the pixel. In addition, the number of categories is known and fixed. Thus, the outputs of semantic segmentation can be easily represented with the output feature maps of the FCNs, and each channel of the output feature maps corresponds to a class. However, in instance segmentation, the prediction target of each pixel is hard to define because instance segmentation also requires to distinguish individual instances, but the number of instances changes in different images.
This poses a major challenge when applying traditional FCNs~\cite{long2015fully} to instance segmentation.

In this work, our core idea is that for an image with $K$ instances, $K$ different mask heads will be dynamically generated, and each mask head will contain the characteristics of its target instance in their filters. As a result, when the mask is applied to an input, it will only fire on the pixels of the instance, thus producing the mask prediction of the instance and distinguishing individual instances. We illustrate the process in Fig.~\ref{fig:mask_heads}. The instance-aware filters are generated by modifying %
an object detector. 
Specifically, %
we add a %
new controller branch to generate the filters for the target instance of each box predicted by the detector, %
as shown in Fig.~\ref{fig:main_figure}. Therefore, the number of the dynamic mask heads is the same as the number of the predicted boxes, which should be the number of the instances in the image if the detector works well. In this work, we build \Ours\ on the popular object detector FCOS~\cite{tian2019fcos} due to its simplicity and flexibility. Also, the elimination of anchor-boxes in FCOS can also save the number of parameters and the amount of computation.

As shown in Fig.~\ref{fig:main_figure}, following FCOS \cite{tian2019fcos}, we make use of the feature maps $\{P_3, P_4, P_5, P_6, P_7\}$ of feature pyramid networks (FPNs) \cite{lin2017feature}, whose down-sampling ratios are $8$, $16$, $32$, $64$ and $128$, respectively. As shown in Fig.~\ref{fig:main_figure}, on each feature level of the FPN, some functional layers (in the dash box) are applied to make instance-aware predictions. For example, the class of the target instance and the dynamically-generated filters for the instance. In this sense, \Ours\ can be viewed as the same as Mask R-CNN, both of which first attend to instances in an image and then predict the pixel-level masks of the instances (\ie, instance-first).

Moreover, recall that Mask R-CNN employs an object detector to predict the bounding-boxes of the instances in the input image. The bounding-boxes are actually the way that Mask R-CNN represents instances. Similarly, \Ours\ employs the instance-aware filters to represent the instances. In other words, instead of encoding the instance information %
with the bounding-boxes, \Ours\ implicitly encodes it with the parameters of the generated dynamic filters, which is much more flexible. For example, the dynamic filters can easily represent the irregular shapes that are hard to be tightly enclosed by a bounding-box (elaborated in Sec.~\ref{sec:gen_filters_encode}). This is one of \Ours's advantages over the previous ROI-based methods.

Besides the detector, as shown in Fig.~\ref{fig:main_figure}, there is also a bottom branch, which provides the feature maps (denoted by $\mF_{bottom}$) that our generated mask heads take as inputs to predict the desired instance mask. The bottom branch aggregates the FPN feature maps $P_3$, $P_4$ and $P_5$. To be specific, $P_4$ and $P_5$ are upsampled to the resolution of $P_3$ with bilinear interpolation and added to $P_3$. After that, four $3 \times 3$ convolutions with $128$ channels are applied. The resolution of the resulting feature maps is the same as $P_3$ (\ie, $\frac{1}{8}$ of the input image resolution). Finally, another convolutional layer is used to reduce the number of the output channels $C_{bottom}$ from $128$ to $8$, resulting in the bottom feature $\mF_{bottom}$. The small output channel reduces the number of the generated parameters. We empirically found that using $C_{bottom} = 8$ can already achieve good performance, and as shown in our experiments, a larger $C_{bottom}$ here (\eg, 16) cannot improve the performance. Even more aggressively, using $C_{bottom} = 1$ only degrades the performance by $\sim1\%$ in mask AP. It is probably because our mask heads only predict relatively simple class-agnostic instance masks and most of the information of an instance has been encoded in the dynamically generated filters.

As mentioned before, the generated filters can also encode the shape and position of the target instance. Since the CNN feature maps do not generally convey the position information, a map of the coordinates needs to be appended to $F_{bottom}$ such that the generated filters are aware of positions.
As the filters are generated with the location-agnostic convolutions, they can only (implicitly) encode the shape and position with the coordinates relative to the location where the filters are generated (\ie, using the coordinate system with the location as the origin). Thus, as shown in Fig.~\ref{fig:main_figure}, $\mF_{bottom}$ is combined with a map of the relative coordinates, which are obtained by transforming all the locations on $\mF_{bottom}$ to the coordinate system with the location generating the filters as the origin. Then, the combination is sent to the mask head to predict the instance mask in the fully convolutional fashion. The relative coordinates provide a strong cue for predicting the instance mask, as shown in our experiments. It is also interesting to note that even if the generated mask heads only take as input the map of the relative coordinates, a modest performance can be obtained as shown in the experiments. This empirically proves that the generated filters indeed encode the shape and position of the target instance. Finally, \sigmoid\ is used as the last layer of the mask head and obtains the mask scores. The mask head only classifies the pixels as the foreground or background. The class of the instance is predicted by the classification head of the detector, as shown in Fig.~\ref{fig:main_figure}.

The resolution of the original mask prediction is same as the resolution of $\mF_{bottom}$, which is $\nicefrac{1}{8}$ of the input image resolution. In order to improve the resolution of instance masks, we use bilinear interpolation to upsample the mask prediction by $2$, resulting in $200 \times 256$ instance masks (if the input image size is $800 \times 1024$). The mask's resolution is much higher than that of Mask R-CNN (only $28 \times 28$ as mentioned before).

\subsection{Network Outputs and Training Targets}

Similar to FCOS, each location on the FPN's feature maps ${P_i}$ either is associated with an instance, thus being a positive sample, or is considered as a negative sample. The associated instance and label for each location are determined as follows.

Let us consider the feature maps $P_i \in \R^{H \times W \times C}$ and let $s$ be its down-sampling ratio. As shown in previous works \cite{tian2019fcos, ren2015faster, he2015spatial}, a location $(x, y)$ on the feature maps can be mapped back onto the input image as $(\floor{\nicefrac{s}{2}} + xs, \floor{\nicefrac{s}{2}} + ys)$. If the mapped location falls in the center region of an instance, the location is considered to be responsible for the instance. Any locations outside the center regions are labeled as negative samples. The center region is defined as the box $(c_x - rs, c_y - rs, c_x + rs, c_y + rs)$, where $(c_x, c_y)$ denotes the mass center of the instance mask, $s$ is the down-sampling ratio of $P_i$ and $r$ is a constant scalar being $1.5$ as in FCOS \cite{tian2019fcos}. As shown in Fig.~\ref{fig:main_figure}, at a location $(x, y)$ on $P_i$, \Ours\ has the following output heads.

\noindent\textbf{Classification Head.} The classification head predicts the class of the instance associated with the location. The ground-truth target is the instance's class $c_i$ or $0$ (\ie, background). As in FCOS, the network predicts a $C$-D vector $\vp_{x, y}$ for the classification and each dimension of $\vp_{x, y}$ corresponds to a binary classifier, where $C$ is the number of categories.

\noindent\textbf{Controller Head.} The controller head, which has the same architecture as the classification head, is used to predict the parameters of the conv.\ filters of the mask head for the instance at the location. The mask head predicts the mask for this particular instance. This is the core contribution of our work. To predict the parameters, we concatenate all the parameters of the filters (\ie, weights and biases) together as an $N$-D vector $\vec{\theta}_{x, y}$, where $N$ is the total number of the parameters. Accordingly, the controller head has $N$ output channels. The mask head is a very compact FCN architecture, which has three $1 \times 1$ convolutions, each having $8$ channels and using \ReLU\ as the activation function except for the last one. No normalization layer such as batch normalization is used here. The last layer has $1$ output channel and uses \sigmoid\ to predict the probability of being foreground. The mask head has $169$ parameters in total ($\#weights = (8 + 2) \times 8 (conv1) + 8 \times 8 (conv2) + 8 \times 1 (conv3)$ and $\#biases = 8 (conv1) + 8 (conv2) + 1 (conv3)$). The masks predicted by the mask heads are supervised with the ground-truth instance masks, which pushes the controller to generate the correct filters.

\noindent\textbf{Box Head.} The box head is the same as that in FCOS, which predicts a 4-D vector encoding the four distances from the location to the four boundaries of the bounding-box of the target instance. Conceptually, \Ours\ can eliminate the box head since \Ours\ needs no ROIs. However, we note that if we make use of box-based NMS, the inference time will be much reduced since we only need to compute the masks for the instances kept after box NMS. Thus, we still predict boxes in \Ours. We would like to highlight that the predicted boxes are \emph{only} used in NMS and do not involve any ROI operations. Moreover, as shown in Table~\ref{table:mask_nms}, the box prediction can be removed if other kinds of NMS are used (\eg, mask NMS \cite{wang2020solov2}). This is fundamentally different from previous ROI-based methods, in which the box prediction is mandatory.

\noindent\textbf{Center-ness Head.} Like FCOS~\cite{tian2019fcos}, at each location, we also predict a center-ness score. The center-ness score depicts how the location deviates from the center of the target instance. In inference, it is used to down-weight the boxes predicted by the locations far from the center as these boxes might be unreliable. The ground-truth center-ness score is defined as
\begin{equation}
    \label{eq:centerness}
    \centerness^* = \sqrt{\frac{ \min(l^*, r^*)}{ \max(l^*, r^*)} %
    \cdot 
        \frac{ \min(t^*, b^*)}{ \max(t^*, b^*)}},
\end{equation}
where $l^*$, $r^*$, $t^*$ and $b^*$ denote the distances from the location to the four boundaries of the ground-truth bounding box. We used the binary cross entropy (BCE) loss to supervise center-ness score as in FCOS.

\begin{figure}[t]
	\centering
	\includegraphics[width=.9\linewidth]{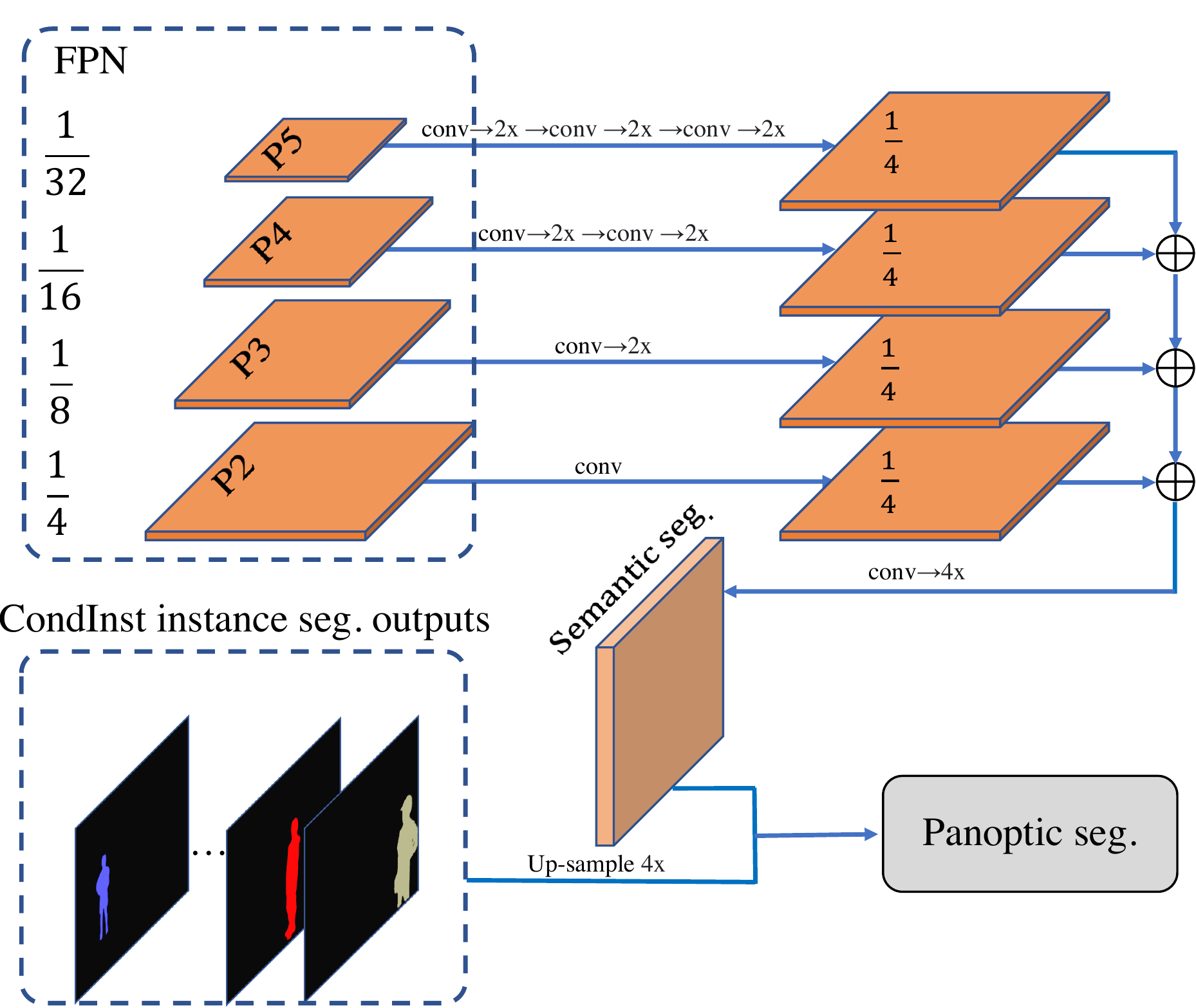}
	\caption{
	Illustration of %
	\Ours\ 
	for 
	panoptic segmentation by attaching a semantic segmentation branch.
	The semantic segmentation branch %
	follows
	\cite{PanoFPN}. 
	Results from the instance segmentation and segmentation segmentation branches are combined together using the same post-processing %
	as in 
	\cite{PanoSeg}. 
	}
	\label{fig:pano_model}
\end{figure}
\noindent\textbf{Semantic Branch for Panoptic Segmentation.} \label{sec:panoptic_seg}
As mentioned before, we can extend \Ours\ to panoptic segmentation by adding a new semantic segmentation branch. For the semantic segmentation branch, we use the structure from Panoptic-FPN \cite{PanoFPN}. To be specific, as shown in Fig.~\ref{fig:pano_model}, the semantic segmentation branch takes as inputs the feature maps $\{P_2, P_3, P_4, P_5\}$ of FPNs. $\{P_3, P_4, P_5\}$ are up-sampled to the same resolution as $P_2$ and the four feature maps are concatenated together. The resolution of $P_2$ is $\nicefrac{1}{4}$ of the input image, which is also the same as the instance masks predicted by \Ours. Then, it is followed by a $1 \times 1$ convolution and $\softmax$ to obtain the semantic segmentation classification scores. The classification scores are trained with the cross-entropy loss. In inference, the semantic segmentation results are merged with the above instance masks to generate the final panoptic segmentation results. The details can be found in Sec.~\ref{sec:inference}.

\subsection{Loss Functions}
Formally, the overall loss function of \Ours\ can be formulated as,
\begin{equation} \label{loss_function_overall}
\begin{aligned}
L_{overall} = L_{fcos} + \lambda L_{mask} + \mu L_{pano},
\end{aligned}
\end{equation}
where $L_{fcos}$ and $L_{mask}$ denote the original loss of FCOS and the loss for instance masks, respectively. $L_{pano}$ (only available in panoptic segmentation) is the loss for the semantic branch of panoptic segmentation. $\lambda$ and $\mu$ being $1$ and $0.5$, respectively, is used to balance these losses. $L_{fcos}$ is the same as in FCOS. Specifically, $L_{fcos}$ includes the classification head, the box regression head and the center-ness head, which are trained with the focal loss~\cite{lin2017focal}, the GIoU loss,
and the binary cross-entropy (BCE) loss, respectively. $L_{mask}$ is defined as,
\begin{equation} \label{loss_function_mask}
\begin{aligned}
&
L_{mask}(\{\vec{\theta}_{x, y}\})  = \frac{1}{N_{\pos}}\sum_{{x, y}}{\mathbbm{1}_{\{c^*_{x, y} > 0\}}L_{dice} \Bigl(\mM_{x, y}, \mM^*_{x, y} \Bigr)},
\end{aligned}
\end{equation}
where $c^*_{x, y}$ is the classification label of location $(x, y)$, which is the class of the instance associated with the location or $0$ (\ie, background) if the location is not associated with any instance. $N_{pos}$ is the number of locations where $c^*_{x, y} > 0$. $\mathbbm{1}_{\{c^*_{x, y} > 0\}}$ is the indicator function, being $1$ if $c^*_{x, y} > 0$ and $0$ otherwise. $\mM^*_{x, y} \in \{0, 1\}^{H \times W}$ is the ground-truth mask of the instance associated with location $(x, y)$, and $\mM_{x, y}$ is the mask predicted by the dynamic mask head of location $(x, y)$. Formally,
\begin{equation}
    \mM_{x, y} = MaskHead(\mat{\tilde{F}}_{x, y}; \vec{\theta}_{x, y}),
\end{equation}
where $\vec{\theta}_{x, y}$ is the generated filters' parameters at location $(x, y)$. $\mat{\tilde{F}}_{x, y} \in \R^{H_{bottom} \times W_{bottom} \times (C_{bottom} + 2)}$ is the combination of $\mat{F}_{bottom}$ and a map of coordinates $\mat{O}_{x, y} \in \R^{H_{bottom} \times W_{bottom} \times 2}$. As described before, $\mat{O}_{x, y}$ is the relative coordinates from all the locations on $\mF_{bottom}$ to $(x, y)$ (\ie, the location where the filters are generated). $MaskHead$ consists of a stack of convolutions with dynamic parameters $\vec{\theta}_{x, y}$.

Moreover, $L_{dice}$ is the  Dice  loss as in \cite{milletari2016v}, which is used to overcome the foreground-background sample imbalance. We do not employ focal loss here as it requires to initialize the biases with a prior probability~\cite{lin2017focal}, which is not trivial if the parameters are dynamically generated. Formally, $L_{dice}$ is defined as
\begin{equation} \label{dice_loss}
\begin{aligned}
&
L_{dice}(\mM, \mM^*)  = 1 - \frac{2 \sum_{i, j}{\mM_{i, j}\mM^*_{i, j}}}{\sum_{i, j}{(\mM_{i, j})^2 + \sum_{i, j}{(\mM^*_{i, j})^2}}},
\end{aligned}
\end{equation}
where $\mM_{i, j}$ or $\mM^*_{i, j}$ denotes the elements of $\mM_{x, y}$ or $\mM^*_{x, y}$, and the subscript $(x, y)$ is omitted for clarification. Note that, in order to compute the loss between the predicted mask $\mM_{x, y}$ and the ground-truth mask $\mM^*_{x, y}$, they need to have the same sizes. As mentioned before, the resolution of the predicted mask $\mM_{x, y}$ is $\nicefrac{1}{4}$ of the ground-truth mask $\mM^*_{x, y}$. Thus, we down-sample $\mM^*_{x, y}$ by $4$ to make the sizes equal. The operation is omitted in Eq.~\eqref{dice_loss} for clarification.

By design, all the positive locations on the feature maps should be used to compute the mask loss. For the images having hundreds of positive locations, the model would consume a large amount of memory. Therefore, in our preliminary version~\cite{tian2020conditional}, the positive locations used in computing the mask loss are limited up to 500 per GPU (\ie, 250 per image %
and we have two images on one GPU%
). If there are more than 500 positive locations, 500 locations will be randomly chosen. In this version, instead of randomly choosing the 500 locations, we first rank the locations by the scores predicted by the FCOS detector, and then choose the locations  with top scores for each instance. As a result, the number of locations per image can be reduced to $64$. This strategy works equally well and further reduces the memory footprint. For instance, 
using this strategy, 
the ResNet-50 based \Ours\ can be trained with 4 1080Ti GPUs.

Moreover, as shown in YOLACT \cite{bolya2019yolact} and BlendMask \cite{chen2020blendmask}, during training, the instance segmentation task can benefit from a joint semantic segmentation task (\ie, using instance masks as semantic labels). Thus, we also conduct experiments with the joint semantic segmentation task, showing improved performance. However, unless explicitly specified, all the experiments in the paper are \emph{without} the semantic segmentation task. If used, the semantic segmentation loss is added to $L_{overall}$.

\subsection{Inference}\label{sec:inference}
\noindent\textbf{Instance Segmentation.} Given an input image, we forward it through the network to obtain the outputs including classification confidence $\vp_{x, y}$, center-ness scores, box prediction $\vt_{x, y}$ and the generated parameters $\vec{\theta}_{x, y}$. We first follow the steps in FCOS to obtain the box detections. Afterwards, box-based NMS with the threshold being $0.6$ is used to remove duplicated detections and then the top $100$ boxes are used to compute masks. Note that each box is also associated with a group of filters generated by the controller. Let us assume that $K$ boxes remain after the NMS, and thus we have $K$ groups of generated filters. The $K$ groups of filters are used to produce $K$ instance-specific mask heads. These instance-specific mask heads are applied, in the fashion of FCNs, to $\mat{\tilde{F}}_{x, y}$ (\ie, the combination of $\mF_{bottom}$ and $\mat{O}_{x, y}$) to predict the masks of the instances. Since the mask head is a very compact network (having three $1 \times 1$ convolutions with $8$ channels and $169$ parameters in total), the overhead of computing masks is extremely small. For example, even with $100$ detections (\ie, the maximum number of detections per image on MS-COCO), only less than $5$ milliseconds in total are spent on the mask heads, which only adds $\sim 10\%$ computational time to the base detector FCOS. In contrast, the mask head of Mask R-CNN has four $3 \times 3$ convolutions with $256$ channels, thus having more than 2.3M parameters and taking longer computational time.

\noindent\textbf{Panoptic Segmentation.} For panoptic segmentation, we follow \cite{PanoFPN} to combine instance and semantic results to obtain the panoptic results. We first rank the instance results from \Ours\ by their confidence scores generated by FCOS. The results with their scores less than $0.45$ are discarded. When overlaps occur between the instance masks, the overlap areas are attributed to the instance with higher score. Moreover, the instance that loses more than 40\% of its total area due to the overlap with other higher-score-instances is discarded. Finally, the semantic results are filled to the areas that are not occupied by any instance.

\begin{table*}[ht!]
    \setlength{\tabcolsep}{6.5pt}
    \centering
	\subfloat[Varying the depth (width $= 8$). \label{varying_depth}]{
		\begin{tabular}{c|c|c|cc|ccc}
			depth & time & AP & AP$_{50}$ & AP$_{75}$ & AP$_{S}$ & AP$_{M}$ & AP$_{L}$ \\
			\Xhline{2\arrayrulewidth}
			1 & \textbf{2.2} & 30.5 & 52.7 & 30.7 & 13.7 & 32.8 & 44.9 \\
			2 & 3.3 & 35.5 & 56.2 & \textbf{37.9} & 17.1 & 38.8 & \textbf{51.2} \\
			3 & 4.5 & \textbf{35.6} & \textbf{56.4} & \textbf{37.9} & \textbf{18.0} & \textbf{38.9} & 50.8  \\
			4 & 5.6 & \textbf{35.6} & 56.3 & 37.8 & 17.3 & 38.9 & 51.0 \\
		\end{tabular}
	}
	\hspace{1em}
	\subfloat[Varying the width (depth $= 3$). \label{varying_width}]
	{
		\begin{tabular}{c|c|c|cc|ccc}
			width & time & AP & AP$_{50}$ & AP$_{75}$ & AP$_{S}$ & AP$_{M}$ & AP$_{L}$ \\
			\Xhline{2\arrayrulewidth}
			2 & \textbf{2.5} & 33.9 & 55.3 & 35.8 & 15.8 & 37.0 & 48.6 \\
			4 & 2.6 & 35.4 & 56.3 & 37.4 & 16.9 & 38.7 & \textbf{51.2} \\
			8 & 4.5 & 35.6 & \textbf{56.4} & 37.9 & \textbf{18.0} & \textbf{39.1} & 50.8  \\
			16 & 4.7 & \textbf{35.7} & 56.1 & \textbf{38.1} & 16.9 & 39.0 & 50.8 \\
		\end{tabular}
	}
	\caption
	{
	Instance segmentation results with different architectures of the mask head on
	the 
	MS-COCO \texttt{val2017} split. ``depth": the number of layers in the mask head. ``width": the number of channels of these layers. ``time": the milliseconds that the mask head takes for processing $100$ instances.
	}
	\label{table:design_choice_mask_head}
\end{table*}

\section{Experiments}
\begin{table}[t]
    \setlength{\tabcolsep}{7.5pt}
    \centering
	\begin{tabular}{c|c|cc|ccc}
		$C_{bottom}$ & AP & AP$_{50}$ & AP$_{75}$ & AP$_{S}$ & AP$_{M}$ & AP$_{L}$ \\
		\Xhline{2\arrayrulewidth}
1 & 34.7 & 56.0 & 36.8 & 16.5 & 37.9 & 50.1 \\
2 & 34.9 & 55.7 & 37.2 & 16.5 & 38.3 & 50.6 \\
4 & 35.5 & 56.3 & 37.5 & 17.8 & 38.7 & 50.7 \\
8 & \textbf{35.6} & \textbf{56.4} & \textbf{37.9} & \textbf{18.0} & \textbf{39.1} & 50.8 \\
16 & 35.4 & 56.0 & 37.5 & 16.9 & 38.7 & \textbf{50.9} \\
	\end{tabular}
	\caption{
	Instance segmentation results by varying the number of channels of the bottom branch's output (\ie, $C_{bottom}$) on the  MS-COCO \texttt{val2017} split. The performance keeps almost the same if $C_{bottom}$ is in a reasonable range, which suggests that \Ours\ is robust to the design choice.}
	\label{table:c_mask}
\end{table}
We evaluate \Ours\ on the large-scale benchmark MS-COCO \cite{lin2014microsoft}. Following the common practice \cite{he2017mask, tian2019fcos, lin2017focal}, our models are trained with split \texttt{train2017} (115K images) and all the ablation experiments are evaluated on split \texttt{val2017} (5K images). Our main results are reported on the \texttt{test}-\texttt{dev} split (20K images).

\subsection{Implementation Details}
Unless specified, we make use of the following implementation details. Following FCOS \cite{tian2019fcos}, ResNet-50  is used as our backbone network and the weights pre-trained on ImageNet \cite{deng2009imagenet} are used to initialize it. For the newly added layers, we initialize them as in \cite{tian2019fcos}. Our models are trained with stochastic gradient descent (SGD) over $8$ V100 GPUs for 90K iterations with the initial learning rate being $0.01$ and a mini-batch of $16$ images. The learning rate is reduced by a factor of $10$ at iteration $60K$ and $80K$, respectively. Weight decay and momentum are set as $0.0001$ and $0.9$, respectively. Following \texttt{Detectron2} \cite{wu2019detectron2}, the input images are resized to have their shorter sides in $[640, 800]$ and their longer sides less or equal to $1333$ during training. Left-right flipping data augmentation is also used during training. When testing, we do not use any data augmentation and only the scale of the shorter side being $800$ is used. The inference time in this work is measured on a single V100 GPU with $1$ image per batch.

\begin{table*}[h]
    \setlength{\tabcolsep}{8.2pt}
	\centering 
	\begin{tabular}{c|c|c|c|cc|ccc|ccc}
		w/ abs.\    coord.   & w/ rel.\  coord. & w/ ${\mF}_{bottom}$ & AP & AP$_{50}$ & AP$_{75}$ & AP$_{S}$ & AP$_{M}$ & AP$_{L}$ & AR$_{1}$ & AR$_{10}$ & AR$_{100}$ \\
		\Xhline{2\arrayrulewidth}
		& & \checkmark & 31.5 & 53.5 & 32.0 & 14.8 & 34.6 & 44.8 & 28.0 & 43.6 & 45.6 \\
		& \checkmark & & 31.3 & 55.0 & 31.9 & 15.6 & 34.1 & 44.3 & 27.1 & 43.3 & 45.6 \\
		\checkmark & & \checkmark & 32.0 & 53.4 & 32.7 & 14.6 & 34.1 & 47.0 & 28.7 & 44.6 & 46.6 \\
		\hline
		& \checkmark & \checkmark & \textbf{35.6} & \textbf{56.4} & \textbf{37.9} & \textbf{18.0} & \textbf{39.1} & \textbf{50.8} & \textbf{30.3} & \textbf{48.7} & \textbf{51.3} \\
	\end{tabular}
	\caption{Ablation study of the input to the mask head on MS-COCO \texttt{val2017} split. As shown in the table, without the relative coordinates, the performance drops significantly from $35.6\%$ to $31.5\%$ in mask AP. Using the absolute coordinates \emph{cannot} improve the performance remarkably. In addition, it is worth noting that if the mask head only takes as inputs the relative coordinates (\ie, no appearance features in this case), \Ours\ also achieves modest performance.}
	\label{table:w_or_wo_offsets}
\end{table*}

\subsection{Architectures of the Mask Head}
In this section, we discuss the design choices of the mask head in \Ours. We show that the performance is not sensitive to the architectures of the mask head. Our baseline is the mask head of three $1 \times 1$ convolutions with $8$ channels (\ie, width $= 8$). As shown in Table~\ref{table:design_choice_mask_head} (3rd row), it achieves $35.6\%$ in mask AP. Next, we first conduct experiments by varying the depth of the mask head. As shown in Table~\ref{varying_depth}, apart from the mask head with depth being $1$, all other mask heads (\ie, depth $= 2, 3$ and $4$) attain similar performance. The mask head with depth being $1$ achieves inferior performance as in this case the mask head is actually a linear mapping, which has overly weak capacity and cannot encode the complex shapes of the instances. Moreover, as shown in Table~\ref{varying_width}, varying the width (\ie, the number of the channels) does not result in a remarkable performance change either as long as the width is in a reasonable range. We also note that our mask head is extremely light-weight as the filters in our mask head are dynamically generated. As shown in Table \ref{table:design_choice_mask_head}, our baseline mask head only takes $4.5$ ms per $100$ instances (the maximum number of instances on MS-COCO), which suggests that our mask head only adds small computational overhead to the base detector. Moreover, our baseline mask head only has $169$ parameters in total. In sharp contrast, the mask head of Mask R-CNN \cite{he2017mask} has more than 2.3M parameters and takes $\sim  \!\!\! 2.5 \times$ computational time ($11.4$ ms per $100$ instances).

\subsection{Design Choices of the Bottom Module}\label{sec:design_choice_mask_branch}
\begin{table}[t]
    \setlength{\tabcolsep}{9pt}
    \centering
	\begin{tabular}{c|c|cc|ccc}
		& AP & AP$_{50}$ & AP$_{75}$ & AP$_{S}$ & AP$_{M}$ & AP$_{L}$ \\
		\Xhline{2\arrayrulewidth}
        $P_3$ & 35.6 & 56.4 & 37.9 & \textbf{18.0} & \textbf{39.1} & 50.8 \\
        $P_2$ & \textbf{36.0} & \textbf{56.6} & \textbf{38.4} & 17.6 & 38.9 & \textbf{51.7} \\
	\end{tabular}
	\caption{
	Instance segmentation results on MS-COCO \texttt{val2017} split by varying the FPN feature level for the bottom module. Using $P_2$ has better performance but it increases the inference latency by about 20\%.}
	\label{table:bottom_p2_p3}
\end{table}
We further investigate the impact of the bottom module. We first change $C_{bottom}$, which is the number of channels of the mask branch's output feature maps (\ie, $\mF_{bottom}$). As shown in Table~\ref{table:c_mask}, as long as $C_{bottom}$ is in a reasonable range (\ie, from $2$ to $16$), the performance keeps almost the same. $C_{bottom} = 8$ is optimal and thus we use $C_{bottom} = 8$ in all other experiments by default.

\begin{figure}[bt]
    \centering
    \includegraphics[width=.95\linewidth]{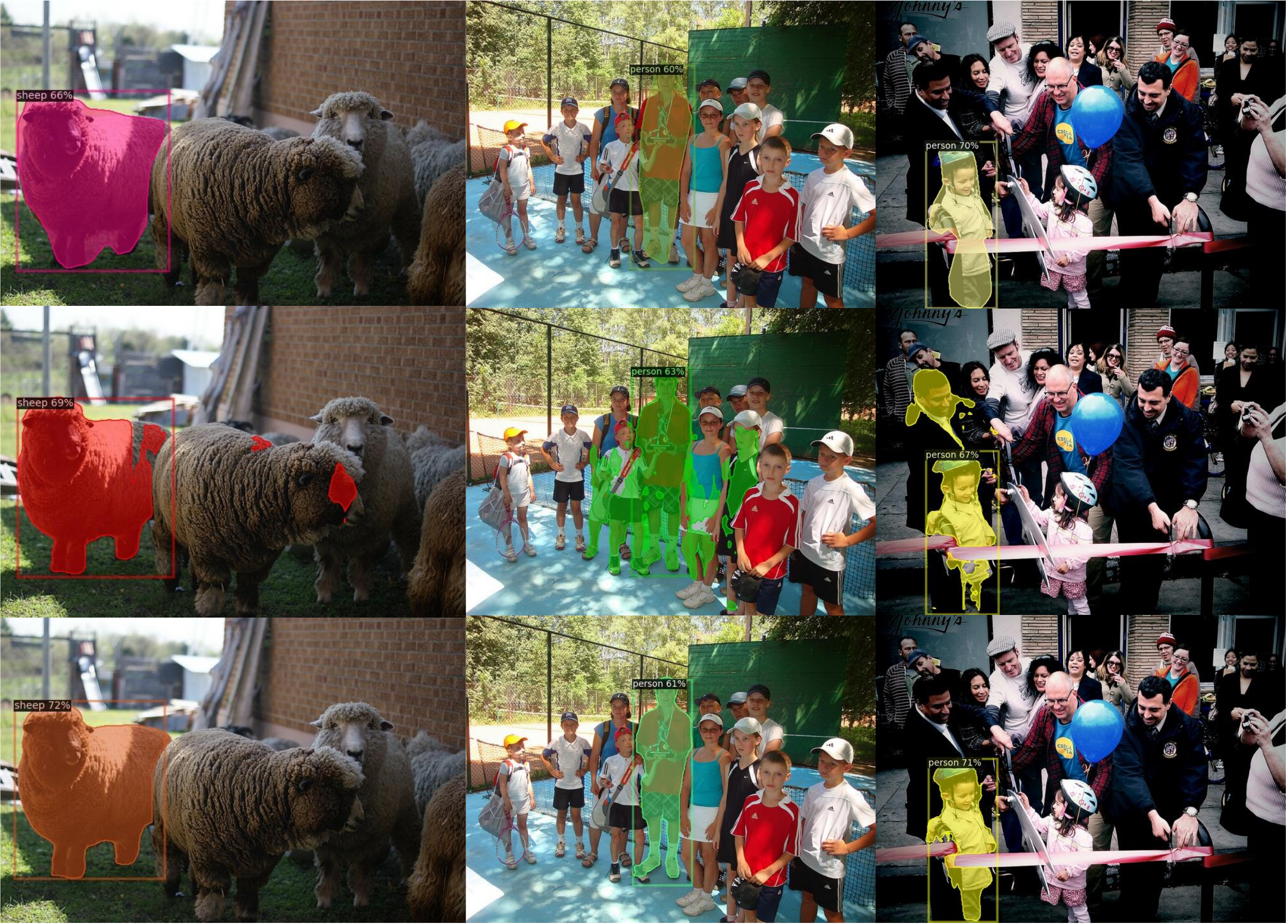}
    \caption{Qualitative results without relative coordinates or bottom features as inputs to the dynamic mask heads. From top to bottom: only with relative coordinates, only with bottom features and with both.
    We can %
    see that the bottom features are crucial to the details of the instance masks, and relative coordinates can help the model distinguish between different instances.}
    \label{fig:feat_coord_vis}
\end{figure}

We conduct experiments by varying the input FPN features of the bottom module. Specifically, we change the FPN feature level from $P_3$ (stride being 8) to $P_2$ (stride being 4) for the bottom module. As shown in Table~\ref{table:bottom_p2_p3}, this can improve the mask AP from $35.6\%$ to $36.0\%$ with 20\% more inference time. Moreover, as mentioned before, before taken as the input of the mask heads, the bottom module's output $\mF_{bottom}$ is concatenated with a map of relative coordinates, which provides a strong cue for the mask prediction. As shown in Table~\ref{table:w_or_wo_offsets} (2nd row), the performance drops significantly if the relative coordinates are removed ($35.6\%$ vs. $31.5\%$). We also experiment with the absolute coordinates, but it cannot largely boost the performance as shown in Table~\ref{table:w_or_wo_offsets} ($32.0\%$). This is understandable because an instance segmentation model should be translation-equivalence. Besides, as shown in Table~\ref{table:w_or_wo_offsets} (2rd row), only using the relative coordinates can also obtain decent performance ($31.3\%$ in mask AP). The qualitative results are shown in \ref{fig:feat_coord_vis}.

\subsection{What %
the Generated Filters Encode?}\label{sec:gen_filters_encode}
It is not
straightforward to 
see what the generated filters encode. However, this can be analyzed by varying the inputs of the dynamic filters and visualizing the changes of the results. As shown in \ref{fig:feat_coord_vis}, it can be noted that if the mask heads only take the relative coordinates as inputs, our model is able to obtain the coarse contour of the instance. This suggests that the generated dynamic filters can attend to the target instance according to the relative coordinates, and it encodes the contour of the target instance. The generated dynamic filters can be also viewed as a representation of a contour. This is different from Mask R-CNN, which attends to a target instance by an axis-aligned RoI produced by Faster R-CNN, \Ours\ encodes the instance's contour into the generated filters. Thus, CondInst can easily represent any shapes including irregular ones, being much more flexible. Moreover, if the bottom features are added, the dynamic filters can produce the details of instance masks. This suggests the generated filters look at the bottom features to obtain the details of the instance masks.

\subsection{How Important to Upsample Mask Predictions?}
\begin{table}[t]
    \setlength{\tabcolsep}{5pt}
    \centering
	\begin{tabular}{c|c|c|cc|ccc}
		factor & resolution & AP & AP$_{50}$ & AP$_{75}$ & AP$_{S}$ & AP$_{M}$ & AP$_{L}$ \\
		\Xhline{2\arrayrulewidth}
		$1$ & $\nicefrac{1}{8}$ & 34.6 & 55.6 & 36.4 & 15.6 & 38.7 & 51.7 \\
		$2$ & $\nicefrac{1}{4}$ & \textbf{35.6} & \textbf{56.4} & \textbf{37.9} & \textbf{18.0} & \textbf{39.1} & \textbf{50.8} \\
		$4$ & $\nicefrac{1}{2}$ &  \textbf{35.6} & 56.2 & 37.7 & 16.9 & 38.8 & \textbf{50.8} \\
	\end{tabular}
	\caption{The instance segmentation results on MS-COCO \texttt{val2017} split by changing the factor used to upsample the mask predictions. ``resolution" denotes the resolution ratio of the mask prediction to the input image. Without the upsampling (\ie, factor $= 1$), the performance drops significantly. The similar results are obtained with ratio $2$ or $4$.} \label{table:upsampling}
\end{table}

As mentioned before, the original mask prediction is upsampled and the upsampling is of great importance to the final performance. We confirm this in the experiment. As shown in Table~\ref{table:upsampling}, without using the upsampling (1st row in the table), in this case \Ours\ can produce the mask prediction with $ \nicefrac{1}{8} $ of the input image resolution, which merely achieves $34.6\%$ in mask AP because most of the details (\eg, the boundary) are lost. If the mask prediction is upsampled by factor $= 2$, the performance can be significantly improved by $1\%$ in mask AP (from $34.6\%$ to $35.6\%$). In particular, the improvement on small objects is large (from $15.6\%$ to $18.0$), which suggests that the upsampling can greatly retain the details of objects. Increasing the upsampling factor to $4$ slightly worsens the performance in some metrics, probably due to the relatively low-quality annotations of MS-COCO. Therefore, we use factor $= 2$ in all other models.

\subsection{\Ours\ without Bounding-box Detection}
Although we still keep the bounding-box detection branch in \Ours, it is conceptually feasible to eliminate it if we make use of the NMS using no bounding-boxes. In this case, all the foreground samples (predicted by the classification head) will be used to compute instance masks, and the duplicated masks will be removed by mask-based NMS. This is confirmed in Table~\ref{table:mask_nms}. As shown in the table, by removing the box branch in inference and using the mask-based NMS, similar performance can be obtained to box-based NMS ($35.6\%$ vs. $35.6\%$ in mask AP). The similar performance of mask and box NMS is probably due to the fact that the instances of MS-COCO are often less dense. Also, although with highly-optimized implementation on GPUs, mask and box NMS can have similar latency, it is worth noting that we need to compute the masks for all the foreground instances before the mask NMS can be applied. The underlying detector FCOS often predicts thousands of foreground instances, and thus it will take significantly longer time to obtain the masks of all the foreground instances. This makes the model with mask NMS significantly slower than the one with box NMS (often more than 2 times slower).

\subsection{Comparisons with State-of-the-art Methods}
We compare \Ours\ against previous state-of-the-art methods on MS-COCO \texttt{test}-\texttt{dev} split. As shown in Table~\ref{table:comparisons_state_of_the_art_methods}, with $1\times$ learning rate schedule (\ie, $90K$ iterations), \Ours\ outperforms the original Mask R-CNN by $0.7\%$ ($35.3\%$ vs. $34.6\%$). \Ours\ also achieves a much faster speed than the original Mask R-CNN ($49$ms vs. $65$ms per image on a single V100 GPU). To our knowledge, it is the first time that a new and simpler instance segmentation method, without any bells and whistles outperforms Mask R-CNN both in accuracy and speed. \Ours\ also obtains better performance ($35.9\%$ vs.\  $35.5\%$) and on-par speed ($49$ms vs.\  $49$ms) than the well-engineered Mask R-CNN in \texttt{Detectron2} (\ie, Mask R-CNN$^*$ in Table~\ref{table:comparisons_state_of_the_art_methods}). Furthermore, with a longer training schedule (\eg, $3\times$) or a stronger backbone (\eg, ResNet-101), a consistent improvement is achieved as well ($37.7\%$  vs.\  $37.5\%$ with ResNet-50 $3\times$ and $39.1\%$ vs.\  $38.8\%$ with ResNet-101 $3\times$). Moreover, as shown in Table~\ref{table:comparisons_state_of_the_art_methods}, with the auxiliary semantic segmentation task, the performance can be boosted from $37.7\%$ to $38.6\%$ (ResNet-50) or from $39.1\%$ to $40.0\%$ (ResNet-101), without increasing the inference time. For fair comparisons, all the inference time here is measured by ourselves on the same hardware with the official code.

\begin{table}[t]
    \setlength{\tabcolsep}{8.7pt}
    \centering
	\begin{tabular}{c|c|cc|ccc}
		NMS & AP & AP$_{50}$ & AP$_{75}$ & AP$_{S}$ & AP$_{M}$ & AP$_{L}$ \\
		\Xhline{2\arrayrulewidth}
		box & \textbf{35.6} & 56.4 & \textbf{37.9} & \textbf{18.0} & \textbf{39.1} & \textbf{50.8} \\
		mask & \textbf{35.6} & \textbf{56.5} & 37.7 & \textbf{18.0} & \textbf{39.1} & 50.7 \\
	\end{tabular}
    \caption{Instance segmentation results with different NMS algorithms. Mask-based NMS can obtain the same overall performance as box-based NMS, which suggests that \Ours\ can eliminate the box detection.} \label{table:mask_nms}
\end{table}
\begin{table*}[h!]
    \setlength{\tabcolsep}{10.7pt}
	\centering
	\begin{tabular}{ l | l |c|c|c|cc|ccc}
	
		method & backbone & aug. & sched. & AP (\%) & AP$_{50}$ & AP$_{75}$ & AP$_{S}$ & AP$_{M}$ & AP$_{L}$ \\
		\Xhline{2\arrayrulewidth}
		Mask R-CNN~\cite{he2017mask} & R-50-FPN & & $1\times$ & 34.6 & \textbf{56.5} & 36.6 & 15.4 & 36.3 & \textbf{49.7} \\
		\textbf{\Ours} & R-50-FPN & & $1\times$ & \textbf{35.3} & 56.4 & \textbf{37.4} & \textbf{18.2} & \textbf{37.8} & 46.7 \\
		\hline
		Mask R-CNN$^*$ & R-50-FPN & \checkmark & $1\times$ & 35.5 & 57.0 & 37.8 & 19.5 & 37.6 & 46.0 \\
		Mask R-CNN$^*$ & R-50-FPN & \checkmark & $3\times$ & 37.5 & 59.3 & 40.2 & \textbf{21.1} & 39.6 & 48.3 \\
		TensorMask~\cite{chen2019tensormask} & R-50-FPN & \checkmark & $6\times$ & 35.4 & 57.2 & 37.3 & 16.3 & 36.8 & 49.3 \\
		BlendMask w/ sem.~\cite{chen2020blendmask} & R-50-FPN & \checkmark & $3\times$ & 37.0 & 58.9 & 39.7 & 17.3 & 39.4 & 52.5\\
		\textbf{\Ours} & R-50-FPN & \checkmark & $1\times$ & 35.9 & 57.0 & 38.2 & 19.0 & 38.6 & 46.7 \\
		\textbf{\Ours} & R-50-FPN & \checkmark & $3\times$ & 37.7 & 58.9 & 40.3 & 20.4 & 40.2 & 48.9 \\
		\textbf{\Ours} w/ sem. & R-50-FPN & \checkmark & $3\times$ & \textbf{38.6} & \textbf{60.2} & \textbf{41.4} & 20.6 & \textbf{41.0} & \textbf{51.1} \\
		\hline
		Mask R-CNN & R-101-FPN & \checkmark & $6\times$ & 38.3 & 61.2 & 40.8 & 18.2 & 40.6 & 54.1 \\
		Mask R-CNN$^*$ & R-101-FPN & \checkmark & $3\times$ & 38.8 & 60.9 & 41.9 & 21.8 & 41.4 & 50.5 \\
		YOLACT-700~\cite{bolya2019yolact} & R-101-FPN & \checkmark & $4.5\times$ & 31.2 & 50.6 & 32.8 & 12.1 & 33.3 & 47.1 \\
		PolarMask~\cite{polarmask} & R-101-FPN & \checkmark & $2\times$ & 32.1 & 53.7 & 33.1 & 14.7 & 33.8 & 45.3 \\
		TensorMask & R-101-FPN & \checkmark & $6\times$ & 37.1 & 59.3 & 39.4 & 17.4 & 39.1 & 51.6 \\
		SOLO~\cite{wang2019solo} & R-101-FPN & \checkmark & $6\times$ & 37.8 & 59.5 & 40.4 & 16.4 & 40.6 & 54.2 \\
		BlendMask$^*$ w/ sem. & R-101-FPN & \checkmark & $3\times$ & 39.6 & 61.6 & 42.6 & \textbf{22.4} & 42.2 & 51.4 \\
		SOLOv2~\cite{wang2020solov2} & R-101-FPN & \checkmark & $6\times$ & 39.7 & 60.7 & \textbf{42.9} & 17.3 & \textbf{42.9} & \textbf{57.4} \\
		\textbf{\Ours} & R-101-FPN & \checkmark & $3\times$ & 39.1 & 60.8 & 41.9 & 21.0 & 41.9 & 50.9 \\
		\textbf{\Ours} w/ sem. & R-101-FPN & \checkmark & $3\times$ & \textbf{40.0} & \textbf{62.0} & \textbf{42.9} & 21.4 & 42.6 & 53.0 \\
		\hline
		\textbf{\Ours} w/ sem. & R-101-BiFPN & \checkmark & $3\times$ & 40.5 & 62.4 & 43.4 & 21.8 & 43.3 & 53.3 \\
		\textbf{\Ours} w/ sem. & 
		DCN-101-BiFPN
		& \checkmark & $3\times$ & \textbf{41.3} & \textbf{63.3} & \textbf{44.4} & \textbf{22.5} & \textbf{43.9} & \textbf{55.2} \\
	\end{tabular}
	\caption{Instance segmentation comparisons with state-of-the-art methods on MS-COCO \texttt{test}-\texttt{dev}. ``Mask R-CNN" is the original Mask R-CNN \cite{he2017mask}. ``Mask R-CNN$^*$" and ``BlendMask$^*$" mean that the models are improved by \texttt{Detectron2} \cite{wu2019detectron2}. ``aug.": using multi-scale data augmentation during training. ``sched.":
	the
	learning rate schedule. 1$\times$ is $90K$ iterations, 2$\times$ is $180K$ iterations and so on. The learning rate is changed as in \cite{he2019rethinking}. ``w/ sem": using the auxiliary semantic segmentation task.
	}
	\label{table:comparisons_state_of_the_art_methods}
\end{table*}

We also compare \Ours\ with the recently-proposed instance segmentation methods. Only with half training iterations, \Ours\ surpasses TensorMask \cite{chen2019tensormask} by a large margin ($37.7\%$ vs.\  $35.4\%$ for ResNet-50 and $39.1\%$ vs.\  $37.1\%$ for ResNet-101). \Ours\ is also $\sim 8\times$ faster than TensorMask ($49$ms vs.\  $380$ms per image on the same GPU) with similar performance ($37.7\%$ vs.\
$37.1\%$). Moreover, \Ours\ outperforms YOLACT-700 \cite{bolya2019yolact} by a large margin with the same backbone ResNet-101 ($40.0\%$ vs.\
$31.2\%$ and both with the auxiliary semantic segmentation task). Moreover, as shown in Fig.~\ref{fig:qualitative}, compared with YOLACT-700 and Mask R-CNN, \Ours\ can preserve more details and produce higher-quality instance segmentation results.

\subsection{Real-time Instance Segmentation with \Ours}
\begin{table}[t]
    \setlength{\tabcolsep}{3.8pt}
	\centering 
	\begin{tabular}{ l |c|c|c|c|cc}
	method & backbone & sched. & FPS & AP & AP$_{50}$ & AP$_{75}$ \\
	\Xhline{2\arrayrulewidth}
	YOLACT-550++~\cite{bolya2019yolact++} & R-50 & $4.5\times$ & 44 & 34.1 & 53.3 & 36.2 \\
	YOLACT-550++ & R-101 & $4.5\times$ & 36 & 34.6 & 53.8 & 36.9 \\
	\hline
	\Ours-RT shtw. & R-50 & $4\times$ & 43 & 36.0 & 57.0 & 38.0 \\
	\Ours-RT shtw. & DLA-34 & $4\times$ & \textbf{47} & 35.8 & 56.5 & 38.0 \\
	\Ours-RT & DLA-34 & $4\times$ & 41 & \textbf{36.3} & \textbf{57.3} & \textbf{38.5} \\
	\end{tabular}
	\caption{The mask AP and inference speed of the real-time CondInst models on the COCO \texttt{test}-\texttt{dev} data. ``shtw.": sharing the conv.\ towers between the classification and box regression branches in FCOS. Both YOLACT++ and \Ours\ use the auxiliary semantic segmentation loss here. As you can see, with the same backbone R-50, \Ours-RT outperforms YOLACT++ by $1.9\%$ AP with almost the same speed. All inference time is measured with a single V100 GPU.} \label{table:real_time_results}
\end{table}

We also present a real-time version of \Ours. Following FCOS~\cite{tian2020fcos}, the $4\times$ conv.\ layers in the classification and box regression towers in FCOS are shared in the real-time models (denoted by ``shtw." in Table~\ref{table:real_time_results}). Moreover, we reduce the input image from a scale of 800 to 512 during testing, and the FPN levels $P_6$ and $P_7$ are removed since there are not many larger objects with the small input images. In order to compensate for the performance loss due to the smaller input size, we use a more aggressive training strategy here. Specifically, the real-time models are trained for $360K$ iterations (\ie, $4\times$) and the shorter side of the input image is randomly chosen from the range 256 to 608 with step 32. Synchronized BatchNorm (SyncBN) is also used during training. In the real-time models, following YOLACT, we enable the extra semantic segmentation loss by default.

The performance and inference speed of these real-time models are shown in Table~\ref{table:real_time_results}. As shown in the table, the R-50 based \Ours-RT outperforms the R-50 based YOLACT++~\cite{bolya2019yolact++} by about $2\%$ AP ($36.0\%$ vs.\  $34.1\%$) and has almost the same speed (43 FPS vs.\  44 FPS). By further using a strong backbone DLA-34~\cite{yu2018deep}, \Ours-RT can achieve 47 FPS with similar performance. Furthermore, if we do not share the classification and box regression towers in FCOS, the performance can be improved to $36.3\%$ AP with slightly longer inference time (41 FPS).

\subsection{Instance Segmentation on Cityscapes}
\begin{table*}[t!]
\setlength{\tabcolsep}{3.63pt}
\centering
\begin{tabular}{ l | l | l |c|cc|cccccccc}
method & backbone & training data & AP [\texttt{val}] & AP & AP$_{50}$ & person & rider & car & truck & bus & train & mcycle & bicycle \\
\Xhline{2\arrayrulewidth}
Mask R-CNN & ResNet-50-FPN & train & 31.5 & 26.2 & 49.9 & 30.5 & 23.7 & 46.9 & 22.8 & 32.2 & 18.6 & 19.1 & 16.0 \\
CondInst & ResNet-50-FPN & train & 33.3 & \textbf{28.6} & \textbf{53.5} & \textbf{31.3} & 23.4 & 51.7 & \textbf{23.4} & \textbf{36.0} & \textbf{27.3} & 19.1 & 16.6 \\
CondInst w/ sem. & ResNet-50-FPN & train & \textbf{33.9} & \textbf{28.6} & 53.1 & \textbf{31.3} & \textbf{24.2} & \textbf{51.9} & 21.2 & 35.9 & 26.5 & \textbf{20.9} & \textbf{17.0} \\
\hline
Mask R-CNN & ResNet-50-FPN & train+COCO & 36.4 & 32.0 & 58.1 & 34.8 & 27.0 & 49.1 & 30.1 & 40.9 & 30.9 & 24.1 & 18.7 \\
CondInst & ResNet-50-FPN & train+COCO & 37.5 & 33.2 & 57.2 & 35.1 & 27.7 & 54.5 & 29.5 & 42.3 & 33.8 & 23.9 & \textbf{18.9} \\
CondInst w/sem. & ResNet-50-FPN & train+COCO & 37.7 & {33.7} & {57.7} & \textbf{35.7} & {28.0} & {54.8} & {29.6} & {41.4} & \textbf{36.3} & \textbf{24.8} & \textbf{18.9} \\
CondInst w/sem. & DCN-101-BiFPN & train+COCO & \textbf{39.3} & \textbf{33.9} & \textbf{58.2} & 35.6 & \textbf{28.1} & \textbf{55.0} & \textbf{32.1} & \textbf{44.2} & 33.6 & 24.5 & 18.6 \\
\hline
CondInst w/sem. & ResNet-50-FPN & train+val+COCO & - & 34.4 & \textbf{59.6} & \textbf{36.4} & 28.4 & 55.3 & 32.6 & 43.3 & 33.9 & 24.8 & \textbf{20.1} \\
CondInst w/sem. & DCN-101-BiFPN & train+val+COCO & - & \textbf{35.1} & 59.0 & 35.9 & \textbf{28.7} & \textbf{55.4} & \textbf{34.4} & \textbf{45.7} & \textbf{35.5} & \textbf{25.5} & 19.6
\end{tabular}
\caption{
Instance segmentation results on Cityscapes \texttt{val} (``AP [\texttt{val}]" column) and \texttt{test} (remaining columns) splits. ``DCN": using deformable convolutions in the backbones. ``+COCO": fine-tuning from the models pre-trained on COCO. ``train+val+COCO": using both \texttt{train} and \texttt{val} splits to train the models evaluated on the \texttt{test} split. ``w/ sem.": using the auxiliary semantic segmentation loss during training as in COCO.}\label{tab:cityscape_result}
\end{table*}

We also conduct the instance segmentation experiments on Cityscapes~\cite{cordts2016cityscapes}. The Cityscapes dataset is designed for the understanding of urban street scenes. For instance segmentation, it has 8 categories, which are person, rider, car, truck, bus, train, motorcycle, and bicycle. It includes 2975, 500 and 1525 images with \texttt{fine} annotations for training, validation and testing, respectively. It also has 20K training images with \texttt{coarse} annotations. Following Mask R-CNN~\cite{he2017mask}, we only use the images with \texttt{fine} annotations to train our models. All images in Cityscapes have the same resolution 2048$\times$1024. The performance on Cityscapes is also measured with the COCO-style mask AP, which are the averaged mask AP over ten IoU thresholds from $0.5$ to $0.95$.

We follow the training details in \texttt{Detectron2}~\cite{wu2019detectron2} to train \Ours\ on Cityscapes. Specifically, the models are trained for 24K iterations with batch size 8 (1 image per GPU). The initial learning rate is $0.01$, which is reduced by a factor of 10 at step 18K. Since Cityscapes has relatively fewer images, following Mask R-CNN, we may initialize the models with the weights pre-trained on the COCO dataset if specified. Moreover, we use multi-scale data augmentation during training and the shorter side of the images is sampled in the range from 800 to 1024 with step 32. In inference, we only use the original image scale 2048$\times$1024. Additionally, in order to preserve more details on Cityscapes, we increase the mask output resolution of \Ours\ from $\nicefrac{1}{4}$ to $\nicefrac{1}{2}$ resolution of the input image.

The results are reported  in Table~\ref{tab:cityscape_result}. As shown in the table, with the same settings, \Ours\ generally outperforms the previous strong baseline Mask R-CNN by more than $1\%$ mask AP in all the experiments. On Cityscapes, the auxiliary semantic segmentation loss can also improve the instance segmentation performance. The results with the loss are denoted by ``w/ sem." in Table~\ref{tab:cityscape_result}. By further using the complementary techniques such as deformable convolutions and BiFPN, the performance can be further boosted as expected. %

\subsection{Experiments on Panoptic Segmentation}

%
%
%
%

%
%
%
%
%
%
%
%
%
%
\begin{figure*}[t]
	\centering

	\includegraphics[width=.986\linewidth]{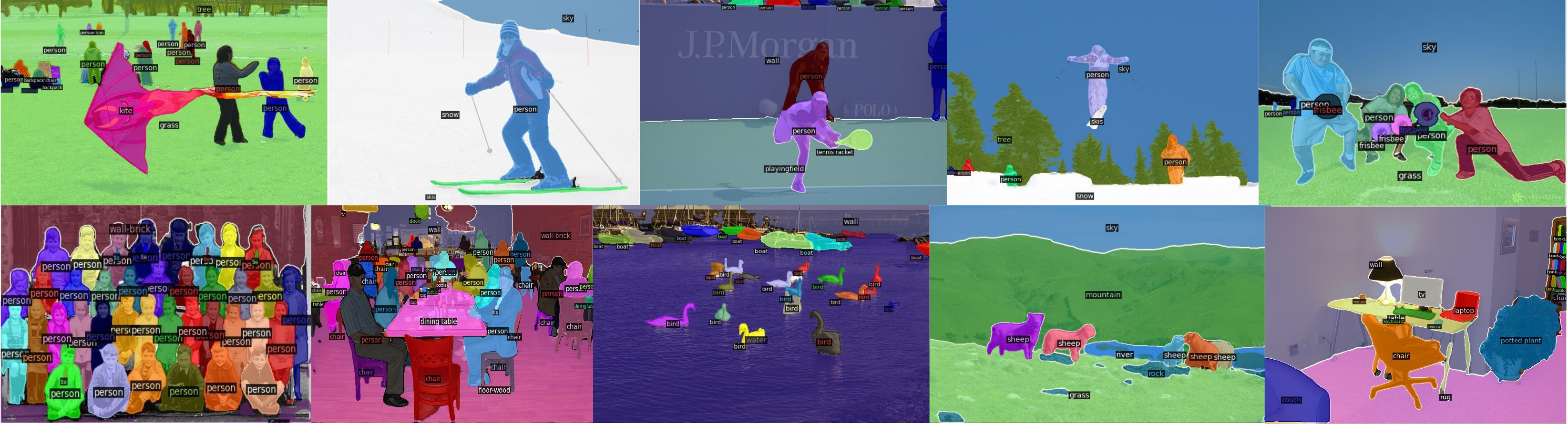}

	\caption{
	Panoptic segmentation results on the COCO dataset (better viewed on screen).
	Color encodes categories and instances. 
	}
	\label{fig:pano_illu}
\end{figure*}

\begin{table}[h!]
        \centering 
        \setlength{\tabcolsep}{2.6pt}
		\begin{tabular}{ l | l |c|ccc}
			method & backbone & sched.\  & PQ & PQ$_{th}$ & PQ$_{st}$  \\
			\Xhline{2\arrayrulewidth}
			\Ours & R-50-FPN & $1\times$ & 42.1 & 50.4 & 29.7  \\
			\hline
			Unifying \cite{li2020unifying} & R-50-FPN & -  & 43.6 & 48.9 & \textbf{35.6}  \\
			
			\Ours & R-50-FPN & $3\times$ & \textbf{44.6} & \textbf{53.0} & 31.8  \\
			\hline
			DeeperLab~\cite{yang2019deeperlab} &  Xception-71~\cite{chollet2017xception} & - & 34.3 & 37.5 & 29.6 \\
			Panoptic-DeepLab~\cite{cheng2019panoptic} & Xception-71 & - & 39.7 & 43.9 & 33.2 \\
			Panoptic-FPN~\cite{PanoFPN}  & R-101-FPN & $3\times$ & 40.9 & 48.3 & 29.7 \\
			AdaptIS~\cite{sofiiuk2019adaptis} & ResNeXt-101 & $1.7\times$ & 42.8 & 50.1 & 31.8 \\
			Aixal DeepLab~\cite{wang2020axial} & Axial-ResNet-L & - & 43.6 & 48.9 & 35.6 \\
			Panoptic-FCN~\cite{li2020fully} & R-101-FPN & $3\times$ &
			45.5& 51.4 & \textbf{36.4} \\
			\textbf{\Ours} & R-101-FPN & $3\times$ & \textbf{46.1} & \textbf{54.7} & 33.2  \\
			\hline
			
			UPSNet~\cite{xiong2019upsnet} & DCN-101-FPN & $3\times$ & 46.6 & 53.2 & 36.7 \\
			
			Panoptic-FCN & DCN-101-FPN & $3\times$ &
			47.1& 53.2 & \textbf{37.8} \\
			
			Unifying & DCN-101-FPN & - & 47.2 & 53.5 & 37.7 \\
			
			\textbf{\Ours} & DCN-101-FPN & $3\times$ & \textbf{47.8}  & \textbf{55.8} & 35.8 \\
		\end{tabular}
    	\caption{Panoptic segmentation 
    	on the COCO \texttt{test}-\texttt{dev} data. All results are with single-model and single-scale testing.
    	Here we report comparisons with state-of-the-art methods %
    	using 
    	various backbones and training schedules ($1\times$ means 90K iterations).
    	\Ours\  achieves the best results among the compared methods.
    	}
    	\label{table:MS-COCO test}
\end{table}

\begin{table}
        \centering 
        \setlength{\tabcolsep}{7pt}
		\begin{tabular}{ l | l |ccc}
			method & backbone & PQ & PQ$_{th}$ & PQ$_{st}$  \\
			\Xhline{2\arrayrulewidth}
			Li \etal \cite{li2018weakly} & - & 53.8 & 42.5 & 62.1\\
			DeeperLab \cite{yang2019deeperlab} & Xception-71 & 56.5 &- &- \\
			Panoptic-FPN \cite{PanoFPN} & R-101-FPN &58.1& 52.0& 62.5\\
			AdaptIS~\cite{sofiiuk2019adaptis} & R-50 & 59.0 & 55.8 & 61.3\\
			UPSNet \cite{xiong2019upsnet} &R-50-FPN & 59.3&54.6&62.7 \\
			Panoptic-DeepLab~\cite{cheng2019panoptic} & R-50 & 59.7 &- &-\\
			Unifying \cite{li2020unifying} & R-50-FPN & 61.4 & 54.7 & 66.3 \\
			Panoptic-FCN~\cite{li2020fully} & R-50-FPN & 61.4 & 54.8 & \textbf{66.6}\\
			
			\hline
			
			\textbf{\Ours} &R-50-FPN &  \textbf{61.7} & \textbf{59.0} & 63.7 \\
		\end{tabular}
    	\caption{Panoptic segmentation 
    	on the Cityscapes \texttt{val} set. All results are with single-model and single-scale with no flipping.
    	Here we report comparisons with state-of-the-art methods.
    	}
    	\label{table:cityscapes_test}
\end{table}

As mentioned before, \Ours\ can be easily extended to panoptic segmentation~\cite{PanoSeg} by %
attaching a 
new semantic segmentation branch depicted in Fig.~\ref{fig:pano_model}. Here, we conduct the panoptic segmentation experiments 
on the COCO 2018 dataset. Unless specified, the training and testing details (\eg, image sizes, the number of iterations and etc.) are the same as in the instance segmentation task on COCO.

Although panoptic segmentation can be viewed as a combination of instance segmentation and semantic segmentation,
there is a discrepancy between the ground-truth annotations of the original instance segmentation and the instance segmentation task in panoptic segmentation. Panoptic segmentation requires that a pixel in the resulting mask has  only  one label. %
Therefore if two instances overlap, the pixels in the overlapped region %
will only be assigned 
to the front instance. However, in the original instance segmentation, the pixels in the overlapped region belong to both instances, and the ground-truth masks are labeled in such a way. Therefore, when we use the instance segmentation framework for panoptic segmentation, the training targets of the instance segmentation
need to be
changed to the instance annotations in panoptic segmentation accordingly. 

We compare our method with %
a few
state-of-the-art panoptic segmentation methods in Table~\ref{table:MS-COCO test}. On the challenging COCO \texttt{test}-\texttt{dev} benchmark, we outperform the previous %
strong baseline Panoptic-FPN~\cite{PanoFPN} by a %
large margin with the same backbone and training schedule (\ie, from $40.9\%$ to $46.1\%$ in PQ with ResNet-101). Moreover, compared to AdaptIS~\cite{sofiiuk2019adaptis}, which shares some similarity with us, the ResNet-101 based \Ours\ achieves dramatically better performance than ResNeXt-101 based AdaptIS ($46.1\%$ vs. $42.8\%$ PQ). This suggests that using the dynamic filters here might be more effective than using FiLM~\cite{perez2018film}. In addition, compared to the recent methods such as ~\cite{li2020unifying} and Panoptic-FCN~\cite{li2020fully}, \Ours\ also outperforms them 
considerably. Some qualitative results are in Fig.~\ref{fig:pano_illu}. We also conduct experiments on the panoptic segmentation task of Cityscapes~\cite{cordts2016cityscapes}, and we follow the training strategy of Panoptic-FPN~\cite{PanoFPN} on this benchmark. Similar to previous works~\cite{li2020fully, li2020unifying, PanoFPN}, we report the results on the Cityscapes \texttt{val} set. As shown in Table~\ref{table:cityscapes_test}, we outperform previous methods on this benchmark as well.

\section{Conclusion}
We have proposed a new and simple instance segmentation framework, termed \Ours.
Unlike previous method such as Mask R-CNN, which employs the mask head with fixed weights, \Ours\ conditions the mask head on instances and dynamically generates the filters of the mask head. This not only reduces the parameters and computational complexity of the mask head, but also eliminates the ROI operations, resulting in a faster and simpler instance segmentation framework. To our knowledge, \Ours\ is the first framework that can outperform Mask R-CNN both in accuracy and speed, without longer training schedules needed. With simple modifications, \Ours\ can be extended to solve panoptic segmentation and achieve state-of-the-art performance on the challenging COCO dataset. We believe that \Ours\ can be a strong alternative for both instance and panoptic segmentation.

\bibliographystyle{ieeetr}
\bibliography{1105}

\vspace{0.25cm} 

\textbf{Authors' photograph and biography not available at the time of publication.}

\input supp.tex

\end{document}

%% file: supp.tex
\clearpage

\appendices

\renewcommand{\theequation}{\thesection\arabic{equation}} 
\renewcommand{\thefigure}{\thesection\arabic{figure}}
\renewcommand{\thetable}{\thesection\arabic{table}}

\renewcommand{\subsubsection}{\subsection}

\section{Visualization of results} 
\begin{figure*}[t]
    \centering
    \includegraphics[width=0.4\textwidth]
          {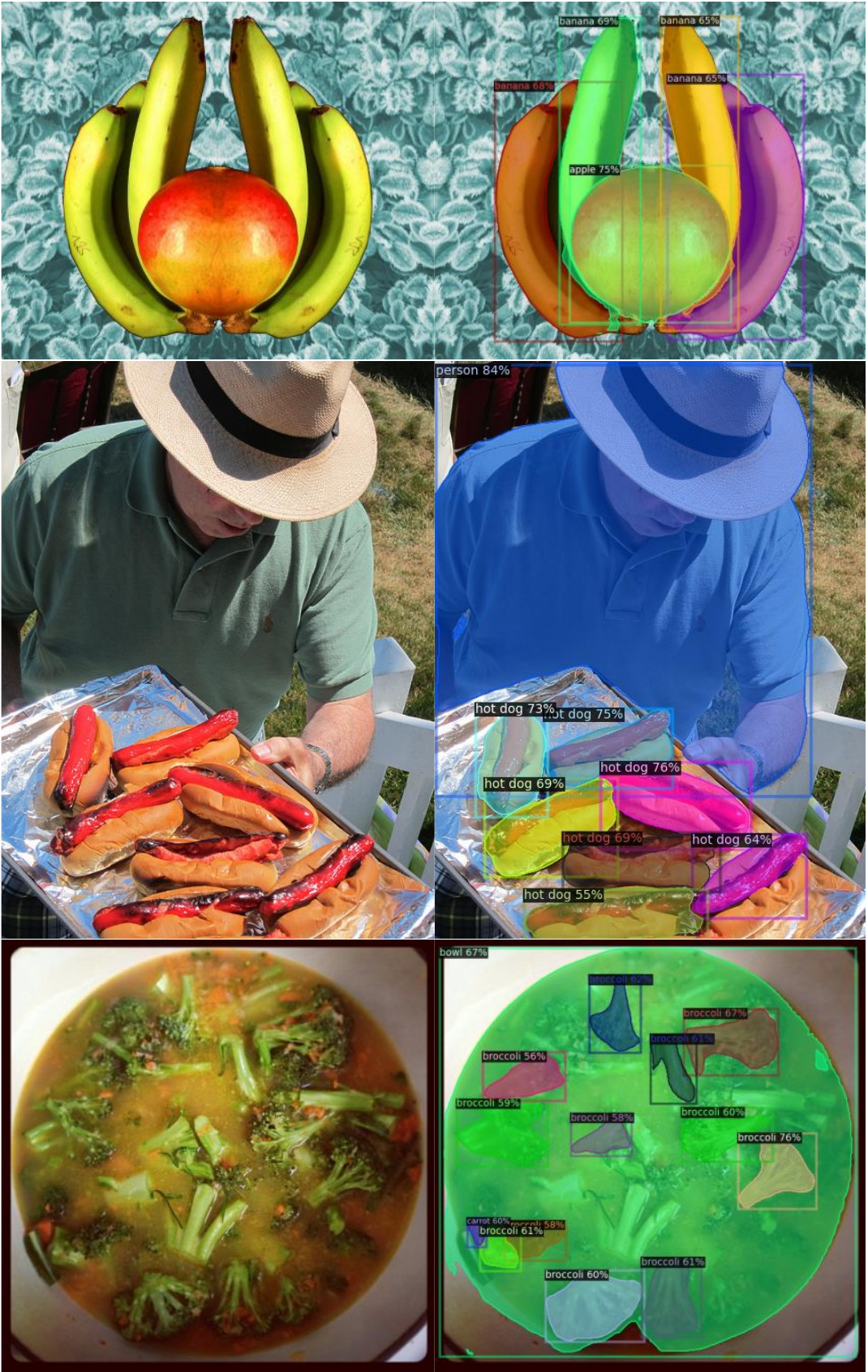}
           \includegraphics[width=0.4\textwidth]{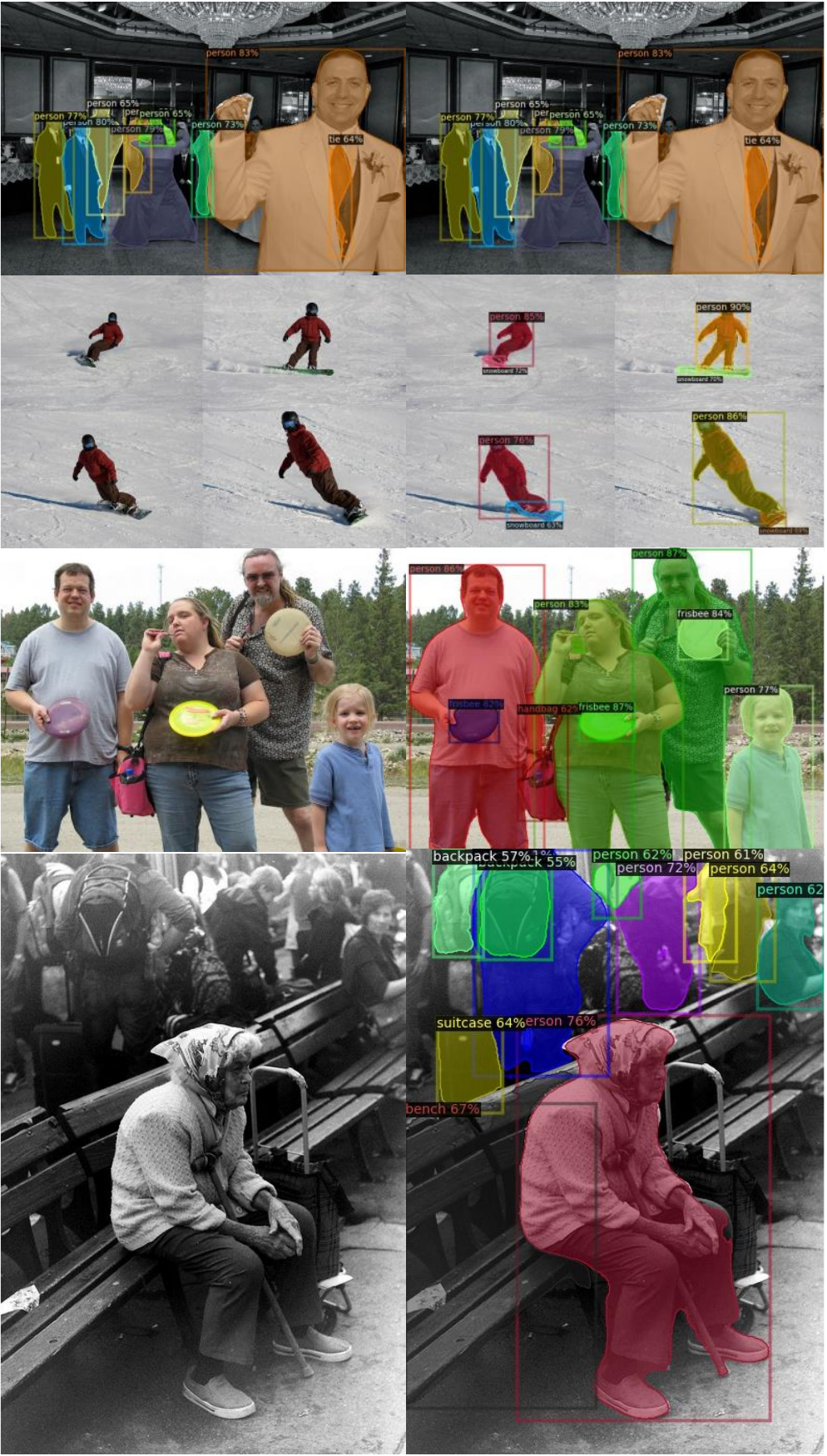}
    \includegraphics[width=0.4\textwidth]{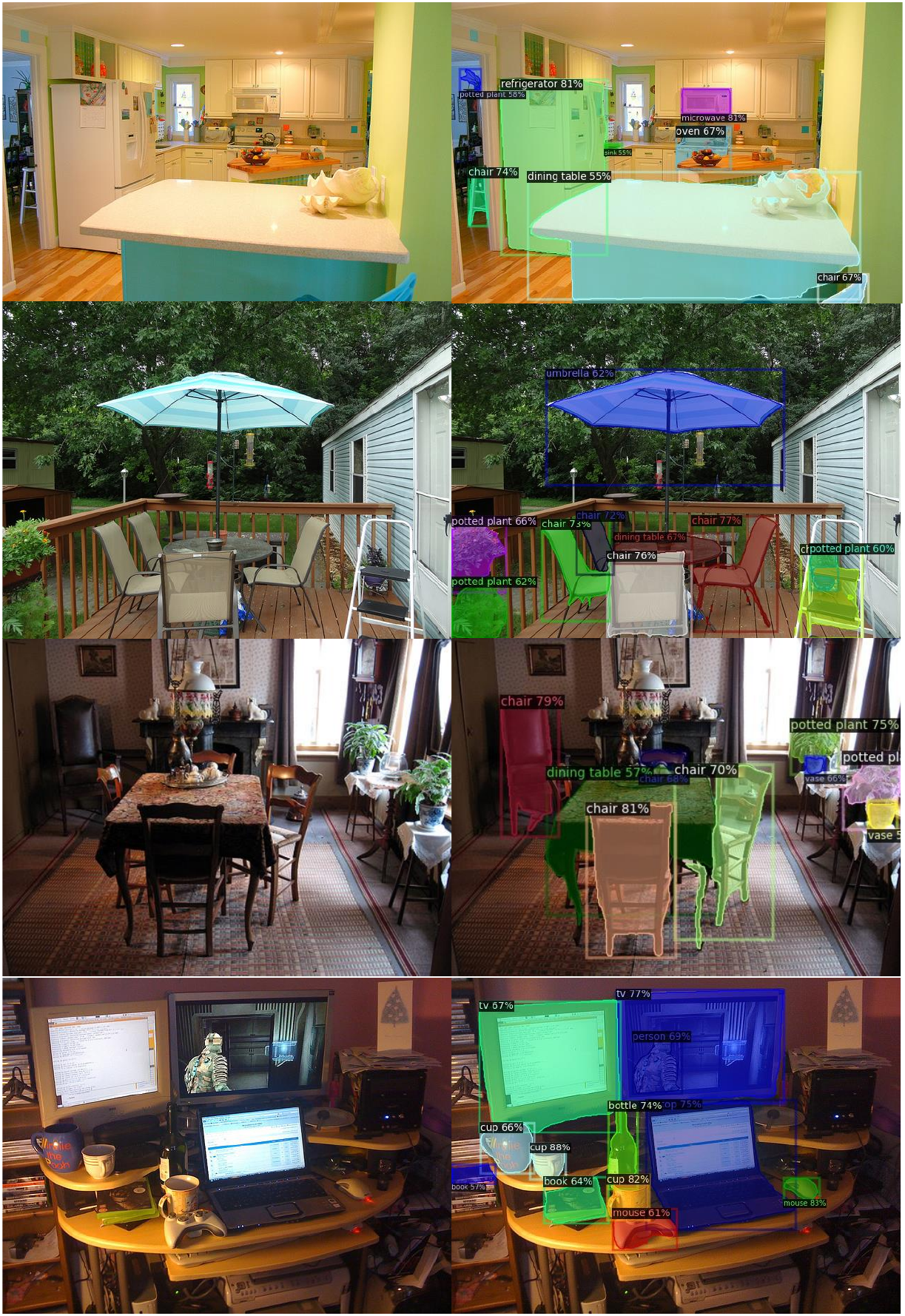}
    \includegraphics[width=0.4\textwidth]{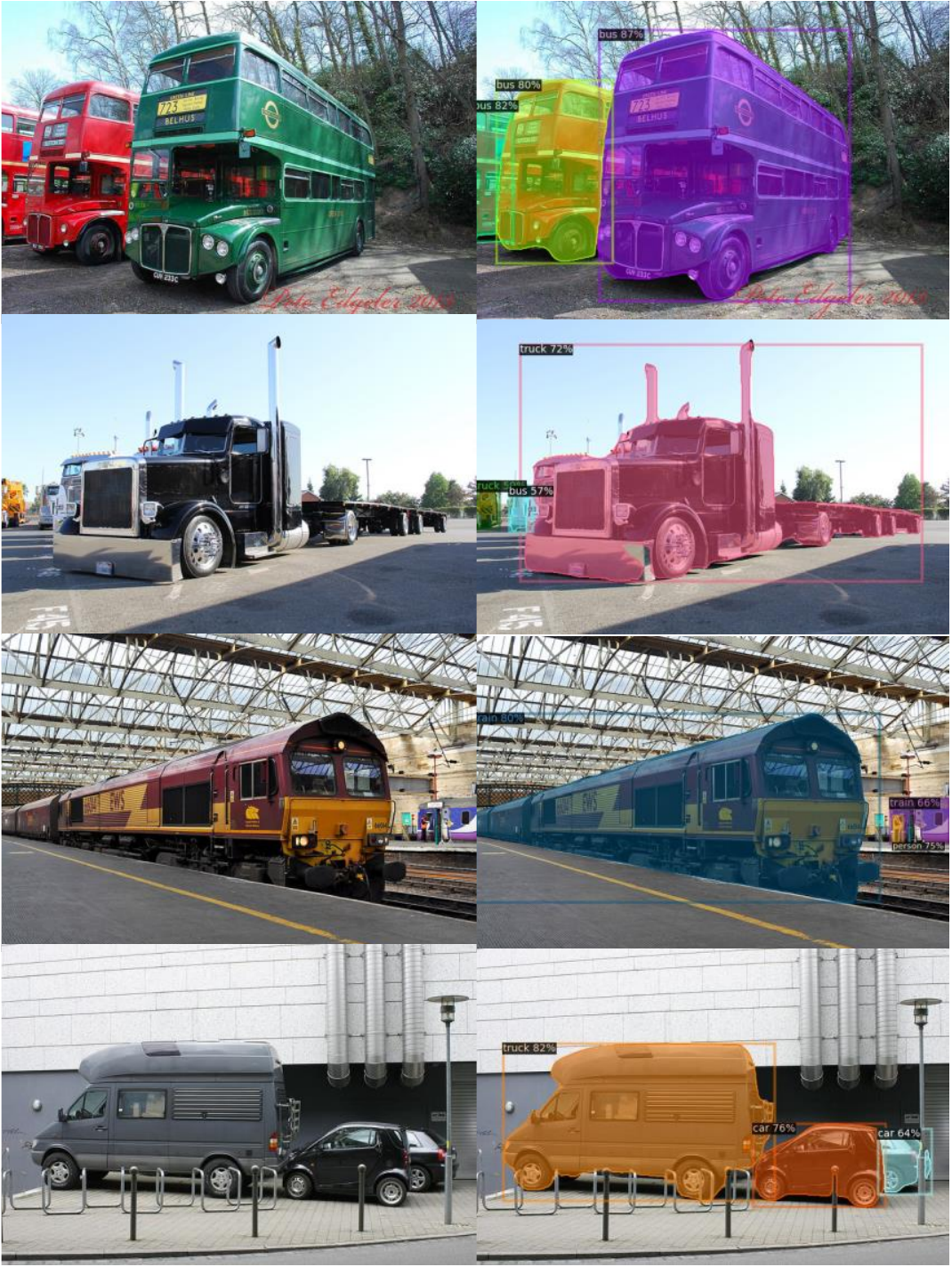}
    \caption{More visualization of instance segmentation results on the COCO dataset (better viewed on screen). Color encodes categories and instances.
    Here the model is ResNet-101-DCN with BiFPN. 
    }
    \label{fig:A1}
\end{figure*}

\begin{figure*}[t]
    \centering
    \includegraphics[width=0.4\textwidth]
          {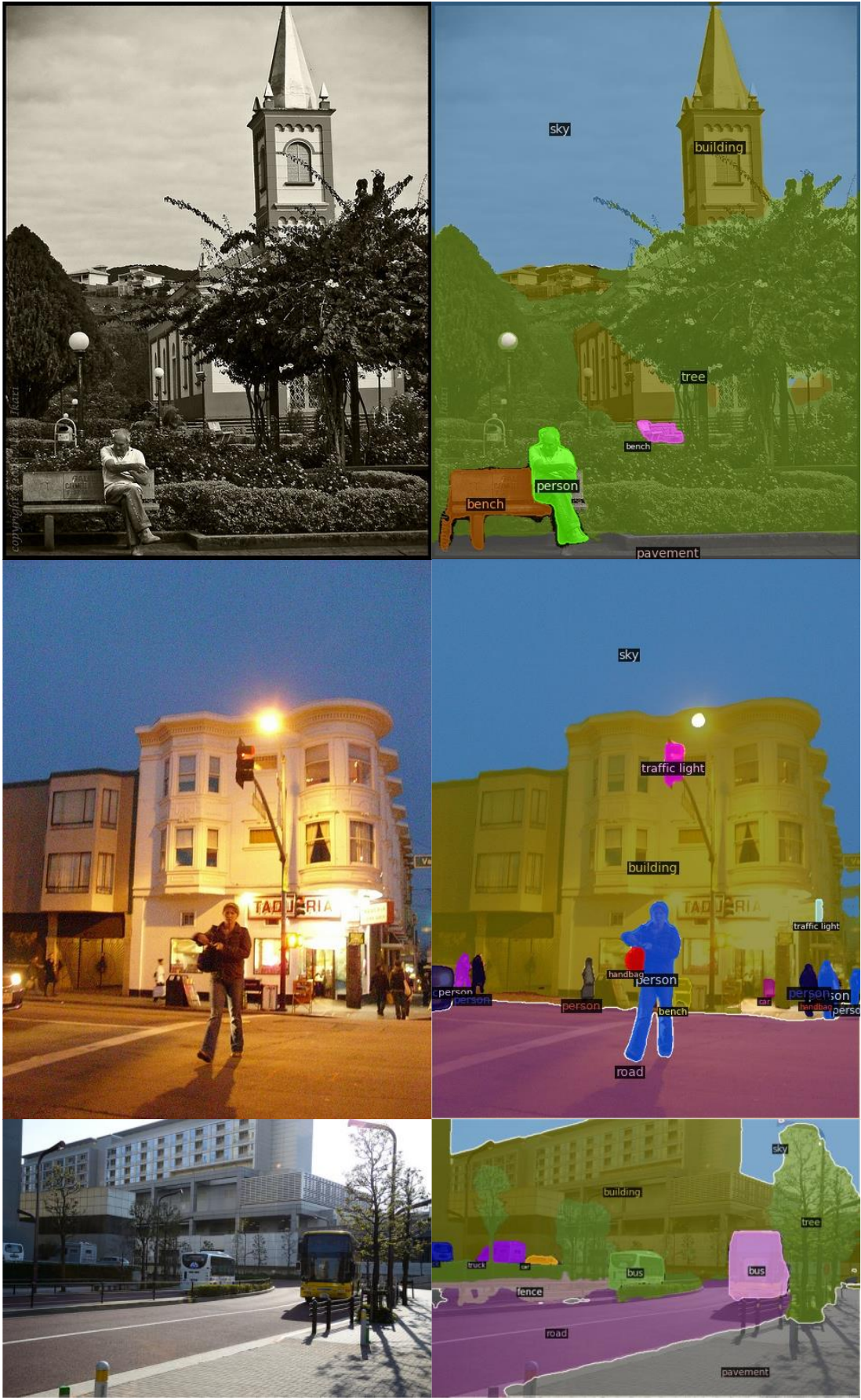}
           \includegraphics[width=0.4\textwidth]{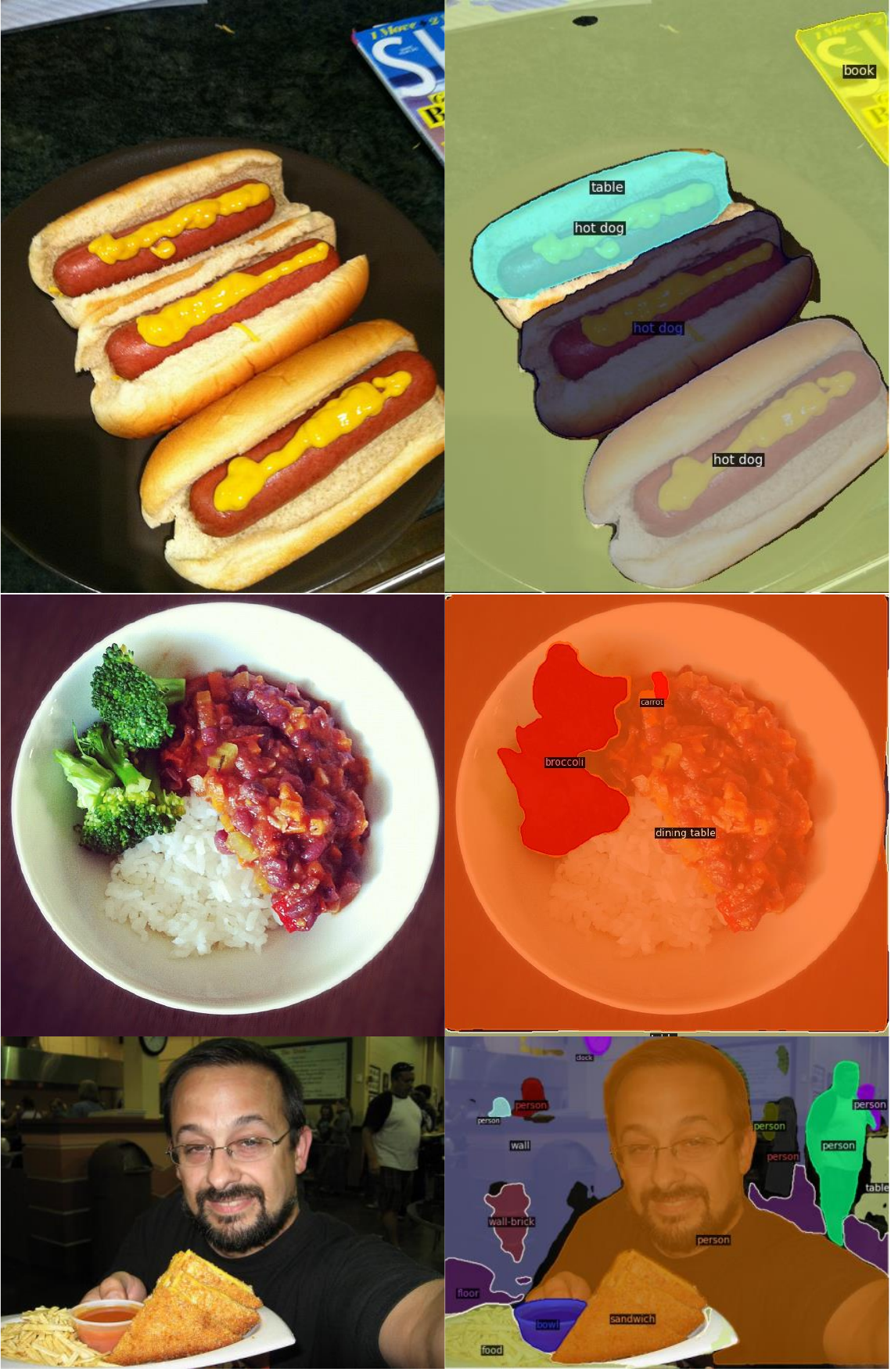}
    \includegraphics[width=0.4\textwidth]{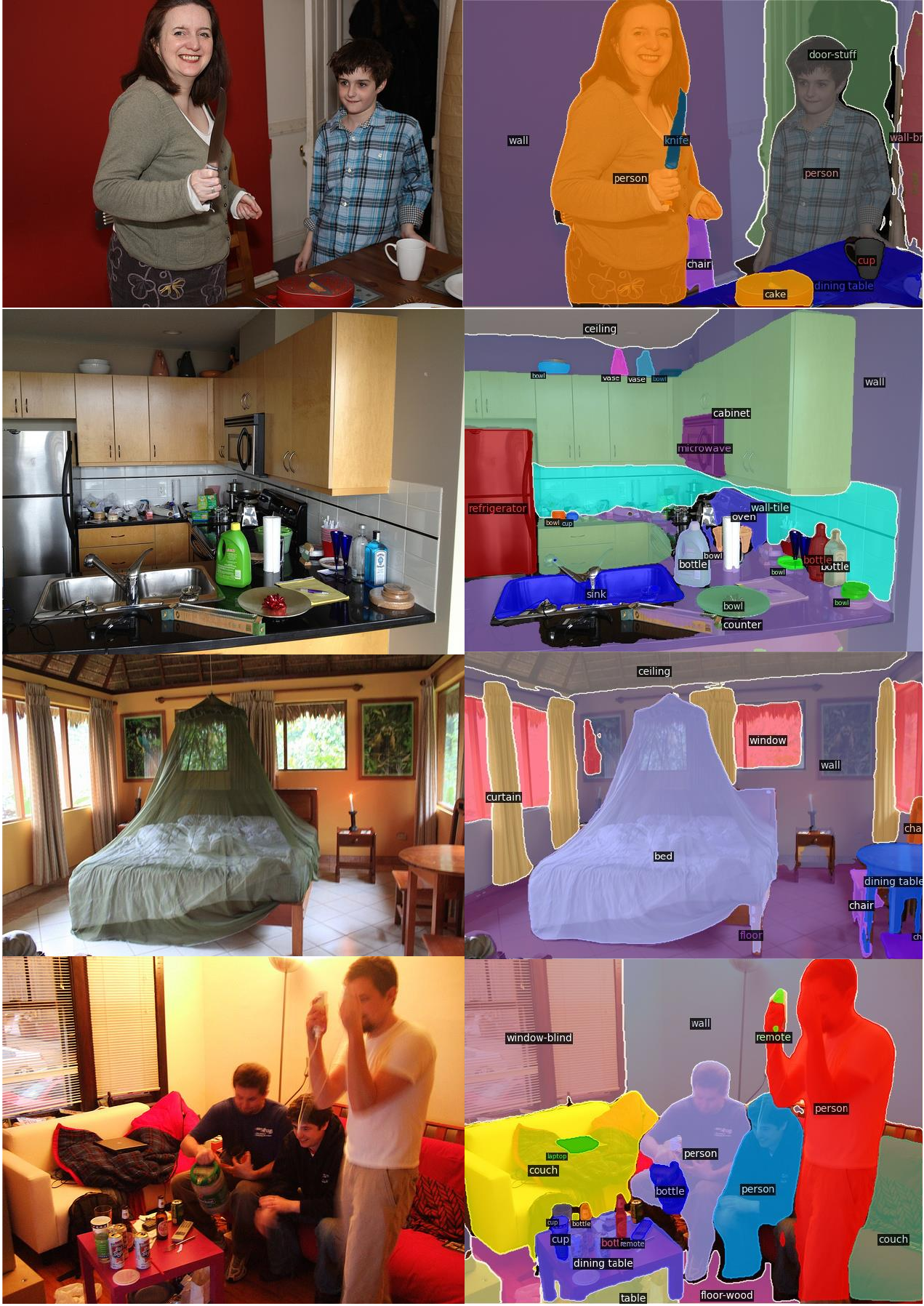}
    \includegraphics[width=0.4\textwidth]{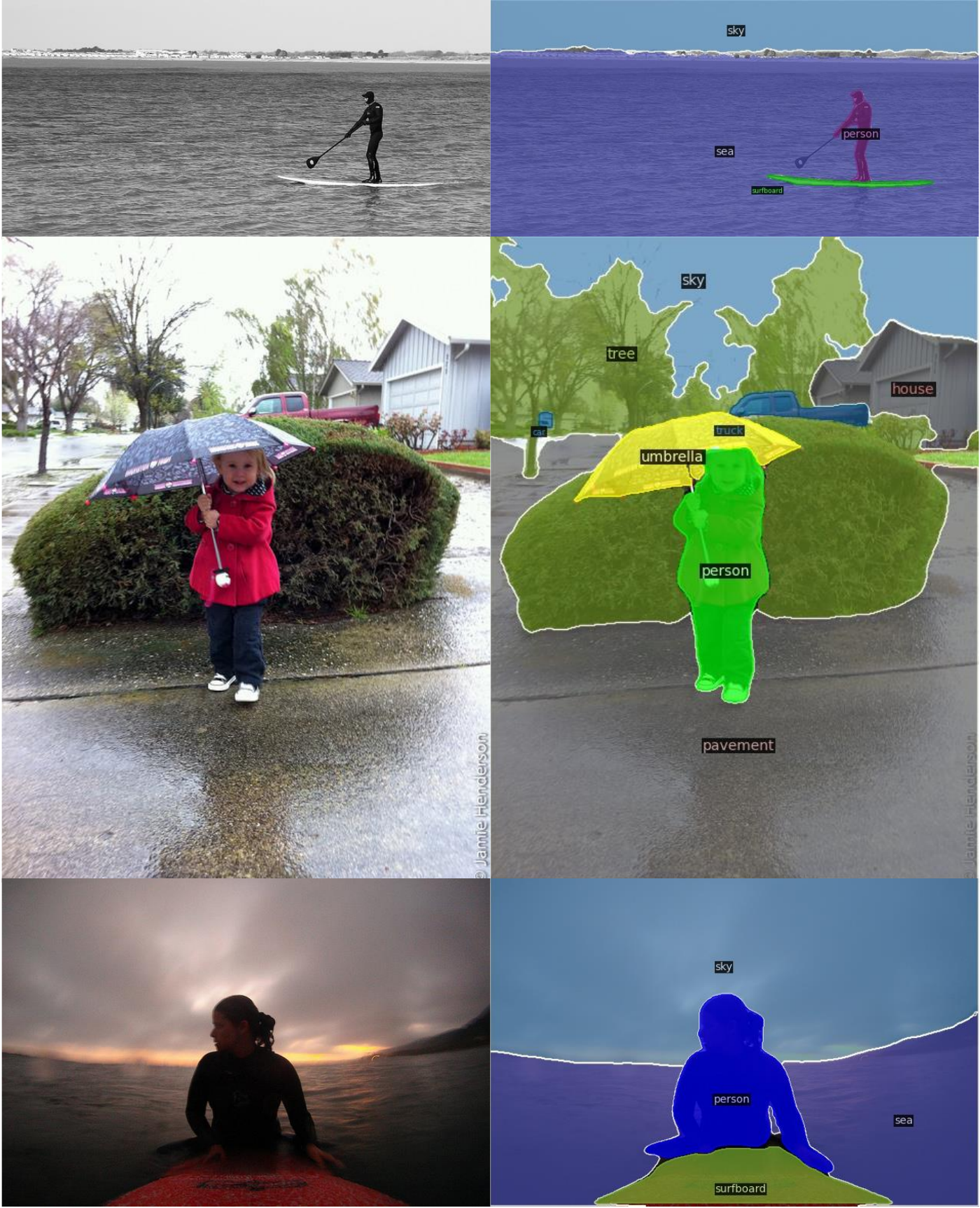}
    \caption{More visualization of panoptic segmentation results on the COCO dataset (better viewed on screen). 
    Here the model is ResNet-101-DCN with standard FPN. 
    }
    \label{fig:A2}
\end{figure*}

Here we provide some visualization results of our model.
Fig.~\ref{fig:A1} and 
Fig.~\ref{fig:A2} show some segmentation results of our model on COCO for instance 
segmentation and panoptic segmentation, respectively. 

\begin{figure*}[t]
    \centering
    \includegraphics[width=0.76\textwidth]{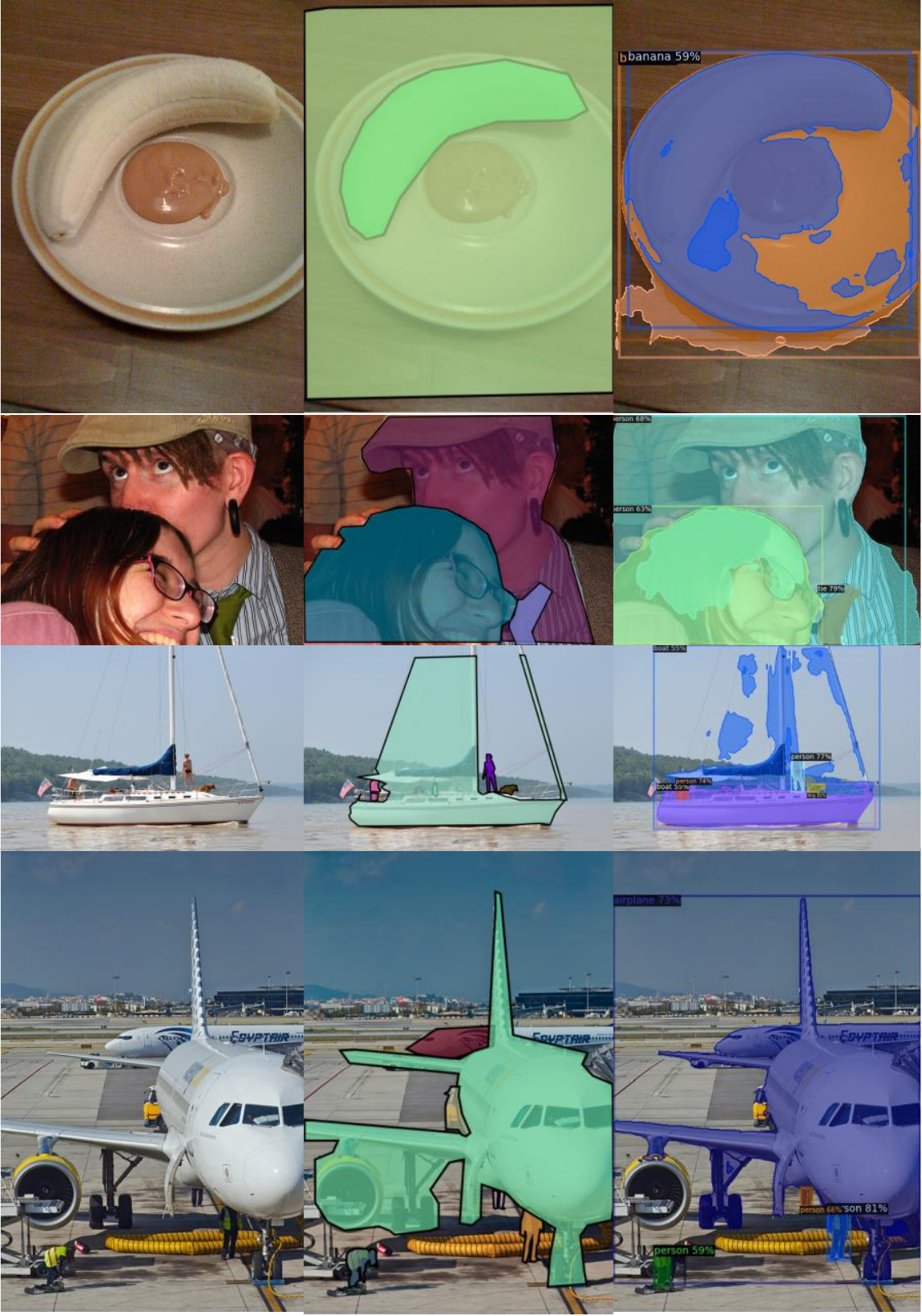}
    \caption{%
    Some 
    instance segmentation results
    that our model does not work very well, 
    on the COCO dataset (better viewed on screen). 
    Left to right: input image, ground-truth labels, model's predictions. 
    In some cases (\textit{e.g.}, the last two examples), the  
    ground-truth annotation is incorrect or noisy. 
    }
    \label{fig:A3}
\end{figure*}

Fig. \ref{fig:A3} show some results that our do not work very well for instance segmentation. In some cases, the COCO annotation is noisy, which may have caused 
confusion for our model. For example, for the third example in Fig. \ref{fig:A3}, 
the sailboat is incorrectly annotated.  Occlusion in the last example also caused challenges. 

\begin{figure*}[t]
    \centering
    \includegraphics[width=0.76\textwidth]{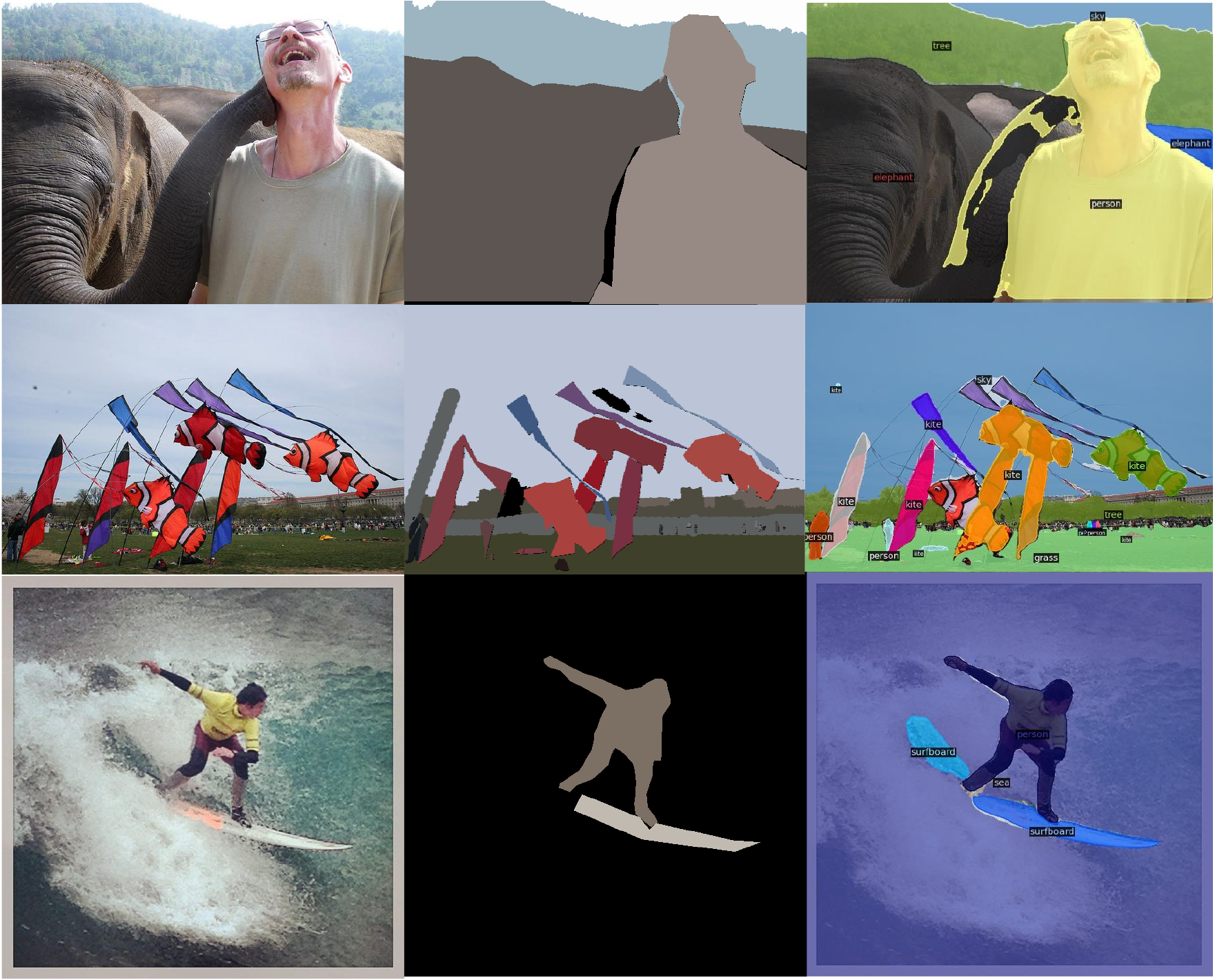}
    \caption{%
    Some 
    panoptic segmentation results
    that our model does not work very well, 
    on the COCO dataset (better viewed on screen). 
    Left to right: input image, ground-truth labels, model's predictions. On those challenging cases, our model makes 
    plausible mistakes. 
    }
    \label{fig:A4}
\end{figure*}

Fig. \ref{fig:A4} shows some panoptic results that our model does not perform well.

